\RequirePackage[svgnames]{xcolor}

\documentclass[11pt,letterpaper]{mystyle}

\usepackage[all]{hypcap}
\usepackage[svgnames,table]{xcolor}
\usepackage[comma,authoryear,compress]{natbib}
\usepackage{hyperref}[citecolor=lightblue]

\hypersetup{
    colorlinks = true,
    citecolor = {YaleBlue},
}

\usepackage{algorithm}
\usepackage{algorithmicx}
\usepackage{algpseudocode}
\usepackage{microtype}
\usepackage{graphicx}
\usepackage{booktabs} %
\usepackage{float}
\usepackage{bigstrut}

\usepackage{amsmath}
\usepackage{amssymb}
\usepackage{mathtools}
\usepackage{amsthm}
\usepackage{mathrsfs}
\usepackage{nicefrac}
\usepackage{dsfont}
\usepackage{enumitem}
\usepackage{subcaption}
\usepackage{cleveref}
\usepackage{bxcoloremoji}

\usepackage{float}

\usepackage[utf8]{inputenc} %
\usepackage[T1]{fontenc}    %
\usepackage{hyperref}       %
\usepackage{url}            %
\usepackage{booktabs}       %
\usepackage{amsfonts}       %
\usepackage{nicefrac}       %
\usepackage{microtype}      %
\usepackage{graphicx}
\usepackage{amssymb}
\usepackage{fdsymbol}
\usepackage{wrapfig}
\usepackage{lipsum}
\usepackage{enumitem}
\usepackage{stackengine}
\usepackage[font=small,labelfont=bf]{caption}
\usepackage{color}
\usepackage{adjustbox}

\usepackage{rotating}
\usepackage{makecell}

\newcommand{\x}{\mathbf{x}}

\definecolor{blanchedalmond}{rgb}{1.0, 0.92, 0.8}
\definecolor{carmine}{rgb}{0.59, 0.0, 0.09}
\definecolor{lightblue}{rgb}{0.22,0.45,0.70}%

\renewcommand{\mathbf}{\boldsymbol}

\makeatletter
\def\Ddots{\mathinner{\mkern1mu\raise\p@
\vbox{\kern7\p@\hbox{.}}\mkern2mu
\raise4\p@\hbox{.}\mkern2mu\raise7\p@\hbox{.}\mkern1mu}}
\makeatother

\definecolor{amaranth}{rgb}{0.9, 0.17, 0.31}
\definecolor{antiquebrass}{rgb}{0.8, 0.58, 0.46}
\definecolor{antiquefuchsia}{rgb}{0.57, 0.36, 0.51}
\definecolor{chromeyellow}{rgb}{0.31, 0.47, 0.26}

\newtcolorbox{AIbox}[2][]{aibox,title=#2,#1}
\definecolor{lightblue}{rgb}{0.22,0.45,0.70}%
\definecolor{Gray}{gray}{0.95}
\definecolor{Cornsilk}{rgb}{1.0, 0.97, 0.86}

\usepackage{amsmath}

\usepackage[all]{hypcap}

\usepackage{multirow}
\usepackage{tabularx}
\usepackage{array}
\usepackage{pifont}
\usepackage{bbm}

\tcbuselibrary{most}

\usepackage{xspace}

\usepackage{cleveref}
\usepackage{listings}
\newcommand{\ourstitle}{FlowBank\xspace}
\newcommand{\ours}{\texttt{{\ourstitle}}\xspace}

\definecolor{colorphase1}{RGB}{178,86,39}
\definecolor{colorphase2}{RGB}{52,117,156}
\definecolor{colorMATH}{RGB}{77,108,125}
\definecolor{colorAMC}{RGB}{113,156,133}
\definecolor{colorMBPP}{RGB}{197,156,123}
\definecolor{colorDROP}{RGB}{152,129,169}
\definecolor{colorMMLU}{RGB}{183,120,118}
\definecolor{boxborder}{RGB}{120,110,170}
\definecolor{boxbg}{RGB}{250,250,250}

\definecolor{darkblue}{rgb}{0, 0, 0.5}
\definecolor{darkred}{rgb}{0.72, 0.22, 0.27}
\definecolor{lightblue}{RGB}{129, 209, 241}
\definecolor{forestgreen}{RGB}{34, 139, 34}

\title{\ourstitle: Query-Adaptive Agentic Workflows Optimization through Precompute-and-Reuse}

\runningtitle{\ourstitle: Query-Adaptive Agentic Workflows Optimization through Precompute-and-Reuse}

\author[1]{Lingzhi Yuan}
\author[1]{Chenghao Deng}
\author[1]{Fangxu Yu}
\author[1]{Souradip Chakraborty}
\author[2]{\protect\\ Mohammad Rostami}
\author[1]{Furong Huang}

\affil[1]{University of Maryland, College Park}
\affil[2]{Amazon}

\correspondingauthor{Furong Huang \href{https://furong-huang.com/}{https://furong-huang.com/}; Email \href{mailto:furongh@umd.edu}{furongh@umd.edu}}

\begin{document}

\begin{abstract}
Large Language Model (LLM)-based multi-agent systems are increasingly powerful, but current agentic workflow optimization paradigms make an unsatisfying trade-off. Task-level methods spend substantial offline compute yet deploy only a single workflow, leaving complementary candidates unused, while query-level methods synthesize a new workflow for each query at substantial inference cost. Our motivating analysis shows that these paradigms are more complementary than competing: workflows discovered during offline search often solve different subsets of queries, and many queries handled by expensive query-level generation can already be solved by cheaper precomputed workflows. This suggests a different objective: rather than searching for one universally best workflow or regenerating a workflow for every instance, we should build a compact bank of reusable, complementary workflows and select among them adaptively at inference time.
Doing so requires solving three coupled problems: generating complementary rather than redundant candidates, compressing them into a small deployable portfolio, and assigning each query to the right workflow under a performance--cost trade-off. To this end, we present \ours, a three-stage framework for portfolio-based agentic workflow optimization. \emph{Diversifying} proposes DiverseFlow to steer search toward under-covered queries and produce a high-coverage candidate pool. \emph{Curating} proposes CuraFlow to compress this pool into a compact portfolio with minimal redundancy. \emph{Matching} casts deployment as edge-value prediction on a query-workflow bipartite graph and routes each incoming query to the portfolio member with the best predicted utility. Across five benchmarks, \ours achieves the highest average score among the evaluated methods while remaining cost-competitive, improving over the strongest automated and handcrafted baselines by 4.26\% and 14.92\% relative, respectively.

\vspace{5pt}
\coloremojicode{1F3E0} \textbf{Project Page}: \href{https://agentic-flowbank.github.io}{https://agentic-flowbank.github.io}
\end{abstract}

\maketitle
\vspace{3mm}

\begin{figure}[!htbp]
\vspace{-1em}
    \centering
    \includegraphics[width=0.85\columnwidth]{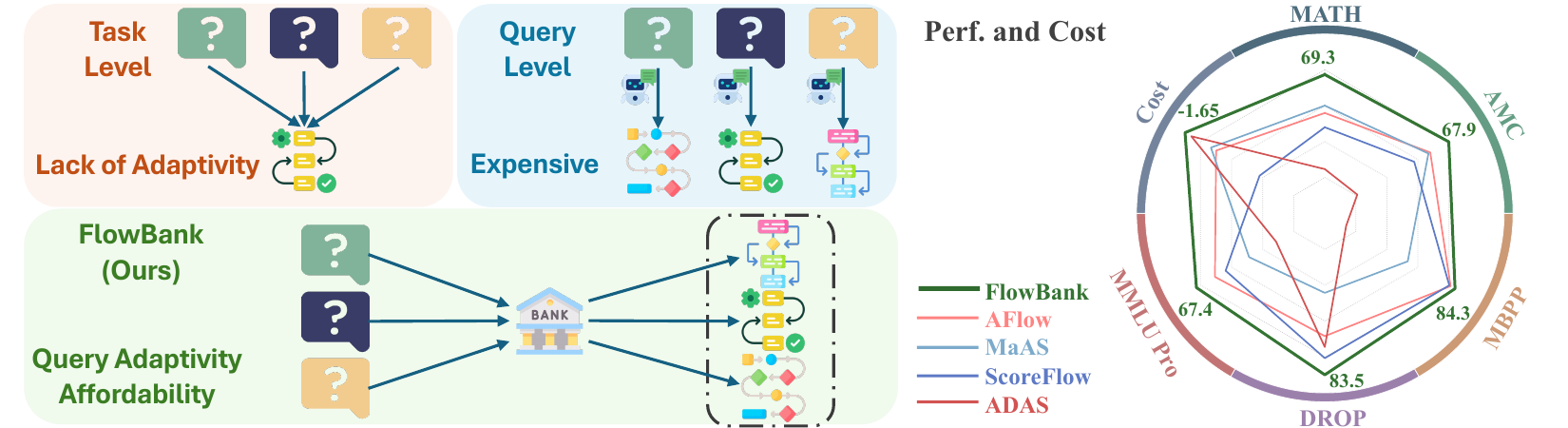}
    \vspace{-0pt}
    \caption{\ours turns workflows from one-shot solutions into reusable assets. \textbf{Left}: Rather than committing to a single universal workflow or generating a new workflow for every query, \ours builds a compact portfolio of complementary workflows offline and assigns each query to the member with the best predicted utility. \textbf{Right}: This portfolio view recovers query-level adaptivity without the full per-query generation cost, yielding a stronger performance--cost trade-off across five benchmarks. (The plotted cost axis is inverted so that higher values correspond to lower actual cost.)}
    \vspace{-4pt}
    \label{fig:teaser}
    \vspace{-8pt}
\end{figure}

\section{Introduction}
\label{sec:intro}

Large Language Models (LLMs) have enabled increasingly capable agentic systems across question answering~\citep{HotpotQA, NQ, WebGPT, MMLU_Pro, yu2024flow, TSRBench}, code generation~\citep{HumanEval, MBPP, DeepSeek-Coder}, web navigation~\citep{WebArena, Mind2web}, and beyond. However, single-agent systems often struggle with intricate real-world requirements, catalyzing a paradigm shift toward multi-agent systems, wherein coordinated ensembles of specialized agents tackle complex tasks~\cite{MetaGPT, MultiAgentDebate, AutoGen}. This raises a central challenge: how to design agentic workflows that are accurate, efficient, and robust across diverse queries~\citep{GPTSwarm, DyLAN}. While many human-designed workflows have been proposed~\cite{SelfConsistency, MultiAgentDebate, MetaGPT, MedPrompt, MultiPersona}, the design space of possible agent topologies, communication patterns, and configurations grows rapidly with system complexity, rendering manual design increasingly untenable and motivating automated workflow optimization~\cite{GPTSwarm, ADAS, AFlow, AgentSquare}.

Existing automated workflow optimization methods largely follow two strategies. Task-level methods run multiple rounds of offline search and then deploy a single universal workflow for all queries, enabling stable and low-cost deployment but enforcing a one-size-fits-all policy~\citep{GPTSwarm, AFlow, ADAS, AgentSquare, Weak-for-strong, MermaidFlow, Do-we-always-need-query-level}. Query-level methods instead synthesize a workflow on the fly for each query, offering stronger adaptivity but pushing substantial reasoning and search cost into inference~\citep{ScoreFlow, MAS-GPT, MaAS, MasRouter, FlowReasoner, MAS2}. The resulting trade-off leaves a natural question: can we recover query-level adaptivity without paying the full cost of per-query workflow generation?

Our motivating analysis reveals two under-exploited forms of workflow complementarity. Within task-level search, workflows that are not selected as the single final winner often solve queries that the chosen workflow misses, so they retain substantial reuse value. Across paradigms, a meaningful portion of the gains achieved by expensive query-level generation can be recovered by cheaper static workflows under query-wise selection, suggesting that part of the online cost is redundant. Taken together, these observations motivate a \ours perspective: rather than commit to one workflow or regenerate a new one for every query, we should precompute a bank of complementary workflows offline and reuse it adaptively at test time.

Turning this perspective into a practical system is challenging. 
\textbf{(C1)} First, \emph{diversified bank construction} requires more than collecting individually strong workflows: existing search procedures optimize candidates largely in isolation, so naively keeping high-scoring outputs tends to produce near-duplicates. 
\textbf{(C2)} Second, \emph{compact subset distilling} is necessary because retaining every discovered workflow increases deployment complexity and selector burden, even though marginal coverage gains quickly diminish. 
\textbf{(C3)} Third, \emph{query-adaptive selecting} remains difficult because the value of a reusable bank is realized only if the system can efficiently match each incoming query to the workflow that offers the best performance--cost trade-off.

To address these challenges, we introduce \ours, a portfolio-based framework organized around three stages: \emph{Diversifying}, \emph{Curating}, and \emph{Matching}. 
\textbf{(S1)} \textbf{Complementarity-oriented Workflow Diversification}: we use DiverseFlow to steer search toward under-covered queries and produce a candidate pool that is complementary rather than merely individually high-scoring.
\textbf{(S2)} \textbf{Coverage-aware Combinatorial Curation}: we use CuraFlow to perform coverage-aware combinatorial search and compress this pool into a compact, high-coverage portfolio. 
\textbf{(S3)} \textbf{Bipartite graph-based Matching}: we cast deployment as edge-value prediction on a query-workflow bipartite graph and assign each incoming query to the portfolio member with the best predicted performance--cost trade-off. 
As shown in \Cref{fig:teaser}, by separating heavy offline workflow discovery from lightweight online selecting, \ours recovers much of the adaptivity of query-level methods while preserving the deployment efficiency of offline methods. 
In this way, \ours turns workflows into reusable assets rather than one-shot design outcomes.

Across five benchmarks spanning mathematical reasoning~\citep{MATH, Easy2hard}, question answering~\citep{MMLU_Pro}, reading comprehension~\citep{DROP}, and code generation~\citep{MBPP}, \ours achieves the highest average score among the evaluated methods while remaining competitive in inference cost, improving over the strongest automated baseline by 4.26\% and the strongest handcrafted baseline by 14.92\% relative, respectively. These results show that selecting from a curated workflow bank is more effective than committing to a single workflow or a single paradigm.

Our main contributions are summarized as follows:
\begin{itemize}[itemsep=0pt, topsep=-1pt, parsep=0pt, partopsep=0pt]
    \item We identify two under-exploited forms of workflow complementarity: discarded task-level workflows retain set-level reuse value, and part of the gain from query-level generation can be recovered through adaptive reuse of cheaper precomputed workflows.
    \item We propose \ours, a three-stage framework that builds a reusable workflow bank through diversified search, coverage-aware curation, and graph-based query-adaptive matching.
    \item We demonstrate that \ours achieves the strongest average performance and competitive performance--cost trade-offs across five benchmarks, validating portfolio-based workflow reuse as a stronger alternative to committing to a single workflow or a single paradigm.
\end{itemize}
\section{Motivating Observations on Workflow Complementarity}
\label{sec:motivation}

\subsection{Problem Formulation}
\label{subsec:preliminaries} 

\paragraph{Agentic Workflows.}
Given a task dataset $D$, an agentic workflow $\omega(\cdot)$ is defined as a computation graph of LLM calls that maps a query $q\in D$ to a prediction $\omega(q)$.
An ideal workflow aims to produce correct answers with lower token usage, balancing accuracy and computational cost.
To quantify this, we evaluate the utility of $\omega$ on a query $q$ using two metrics:
\textbf{(1)} the performance $e(\omega, q)  \in [0, 1]$, measured by a task-specific evaluation function,
and \textbf{(2)} the cost $c(\omega, q) > 0$, defined as the total token consumption incurred by $\omega$ while processing $q$.

\paragraph{Agentic Workflow Optimization.}
Agentic workflow optimization aims to maximize the performance of a workflow on a training dataset $D_{\text{train}}\subset D$.
Existing agentic workflow optimization methods generally fall into two paradigms.

\textit{Task-level} optimization produces a static workflow for all queries in $D_{\text{train}}$ to maximize the overall performance, i.e.,
$\omega_{\text{stat}} = \mathop{\arg\max}_{\omega}\sum_{q\in D_{\text{train}}} e(\omega, q).$
This workflow is chosen through multiple candidates discovered through iterative exploration of the workflow space.
The per-query cost is affordable in deployment, as tokens are only consumed during workflow execution.
But using a single workflow for all queries may limit deployment performance due to the lack of query adaptation.
    
\textit{Query-level} optimization uses a meta-generator $G_{\phi}$ to design a dynamic workflow $\omega_{\mathrm{dyn}}^q = G_{\phi}(q)$ for each query $q \in D$. 
The optimization objective is to maximize the average performance of dynamic workflow on $D_{\text{train}}$ through adjusting the parameters of $G_{\phi}$, i.e.,
$\phi^*=\mathop{\arg\max}_{\phi}\sum_{q\in D_{\text{train}}} e(G_\phi(q), q)$.
The per-query cost of this paradigm includes both workflow generation and workflow execution.
This paradigm enjoys query adaptation by workflow customization but the high inference cost can make large-scale deployment expensive.

\subsection{Motivating Analysis}
\label{subsec:motivation_analysis}

\paragraph{Potential of Reusing Workflows through Task-level Searching.}
\begin{wrapfigure}[14]{r}{0.40\columnwidth}
    \vspace{-1.5\baselineskip}
    \centering
    \includegraphics[width=0.38\columnwidth]{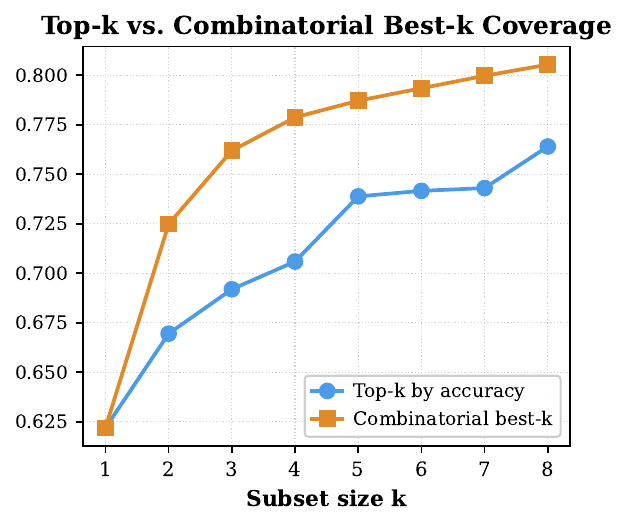}
    \vspace{-6pt}
    \captionsetup{hypcap=false}
    \caption{Coverage on MATH for workflow sets built from AFlow candidates.}
    \label{fig:sec_2_tasklevel}
    \vspace{-0.3\baselineskip}
\end{wrapfigure}
Task-level workflow optimization explores many candidate workflows, but deployment ultimately uses only one. This raises a natural question: do the remaining workflows still provide reusable value as a set? To quantify this on a finite dataset $D$, we define the \emph{coverage} of a workflow set
\vspace{-2pt}
\begin{equation}
\mathrm{Coverage}(\Omega) = \frac{1}{|D|} \sum_{q\in D} \max_{\omega \in \Omega} e(\omega, q).
\end{equation}
\vspace{-1pt}
We instantiate this analysis with AFlow~\citep{AFlow}, a representative task-level workflow optimizer that performs offline Monte Carlo tree search and ultimately deploys a fixed workflow. Specifically, we study the 20 candidate workflows produced by 20 rounds of AFlow optimization on MATH~\citep{MATH}. 
For each set size $k \in \{1,\dots,20\}$, we first select the top-$k$ workflows ranked by individual accuracy. 
As shown by the blue curve in Fig.~\ref{fig:sec_2_tasklevel}, the coverage of this set increases steadily with $k$, indicating that unselected workflows for final deployment still retain reusable value when combined as a set. We then compare it with the combinatorially best size-$k$ set, obtained by exhaustive search over all $\binom{20}{k}$ subsets, shown by the orange curve. The consistent gap between the two curves shows that accuracy-based construction is suboptimal for set coverage. 
This suggests a workflow pool optimized for single-workflow accuracy may not be the best one for set-level reuse.

\paragraph{Potential of Reducing Redundant Workflow Generation for Query Adaptation.}
\begin{wrapfigure}[15]{r}{0.40\columnwidth}
    \vspace{-0.4\baselineskip}
    \centering
    \includegraphics[width=0.38\columnwidth]{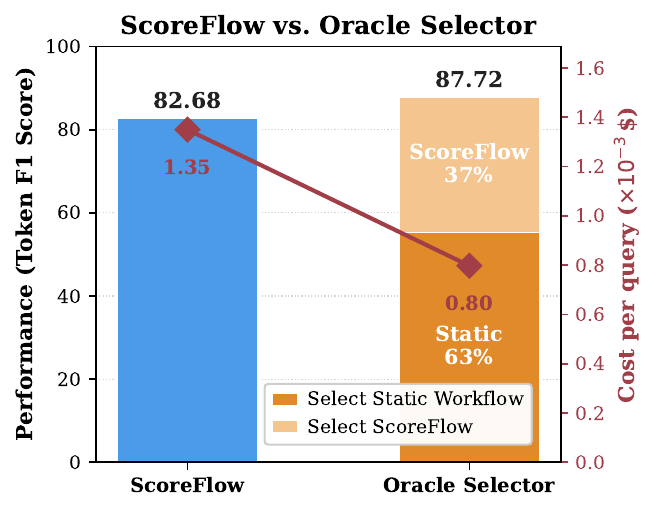}
    \vspace{-3pt}
    \captionsetup{hypcap=false}
    \caption{Performance--cost comparison on DROP between ScoreFlow and an oracle selector.}
    \label{fig:sec_2_cross_para}
    \vspace{-0.3\baselineskip}
\end{wrapfigure}
To examine whether the extra inference cost for query adaptation is always necessary, we use ScoreFlow~\citep{ScoreFlow} as a representative query-level method, which uses DPO to train a meta-generator and designs a customized workflow for each query at inference time. 
We combine it with the best static workflow by AFlow. 
For each query, an oracle chooses the workflow with higher performance, breaking ties in favor of the one with lower cost. 
As shown in Fig.~\ref{fig:sec_2_cross_para}, a substantial fraction of the queries solved by ScoreFlow can be handled by the cheaper static workflow. 
This suggests that token cost in query-level method can be reduced by diverting queries smartly. 
The oracle's superior performance--cost trade-off further highlights the potential of integrating static and query-level workflows within a unified framework rather than treating them as competing paradigms.

These observations reveal substantial potential in workflow complementarity: reusing multiple static workflows can improve coverage, and combining static and query-level workflows can further improve the performance--cost trade-off. Realizing this potential, however, requires addressing two challenges. \textbf{(1)} \emph{Workflow pool construction}: traditional task-level methods optimize workflows for individual accuracy, which is suboptimal for building a high-coverage workflow pool. \textbf{(2)} \emph{Query-adaptive selection}: the gains can be realized only if the system chooses the right workflow for each query at inference time efficiently. 
These challenges directly motivate our design of \ours.
\section{Methodology}
\label{sec:method}

To address the above challenges, \ours adopts a portfolio-based methodology for workflow deployment. Rather than searching for a single universally best workflow, we aim to build, curate, and deploy a compact set of complementary workflows. Concretely, as shown in \Cref{fig:pipeline}, \ours consists of three stages. In \emph{Diversifying}, DiverseFlow explores the workflow search space to generate a raw candidate pool with broad query coverage. In \emph{Curating}, CuraFlow distills this pool into a compact portfolio that preserves most attainable coverage while removing redundant workflows. In \emph{Matching}, we cast workflow assignment as edge-value prediction on a query-workflow bipartite graph and select the portfolio member with the best predicted utility.

\subsection{Diversifying via DiverseFlow}
\label{subsec:diverseflow}
\paragraph{Workflow Candidate Pool Generation.}
We propose DiverseFlow, built on Monte Carlo tree search (MCTS), to generate a workflow candidate pool with broad coverage.
It inherits a four-step search structure for each round in \citep{AFlow}.
\textbf{(1)} Among the discovered workflows, one is sampled as the parent workflow.
\textbf{(2)} Conditioned on the parent workflow and tree-structured optimization experience, the optimizer LLM proposes a new workflow in the \textit{code} space.
\textbf{(3)} The new workflow is then evaluated on $D_{\text{train}}$ and added to the candidate pool.
\textbf{(4)} The optimization experience of the current round is tagged as Success/Failure based on whether it outperforms the parent and is then attached to the tree.

\begin{figure}[t] 
    \centering 
    \includegraphics[width=1.0\textwidth]{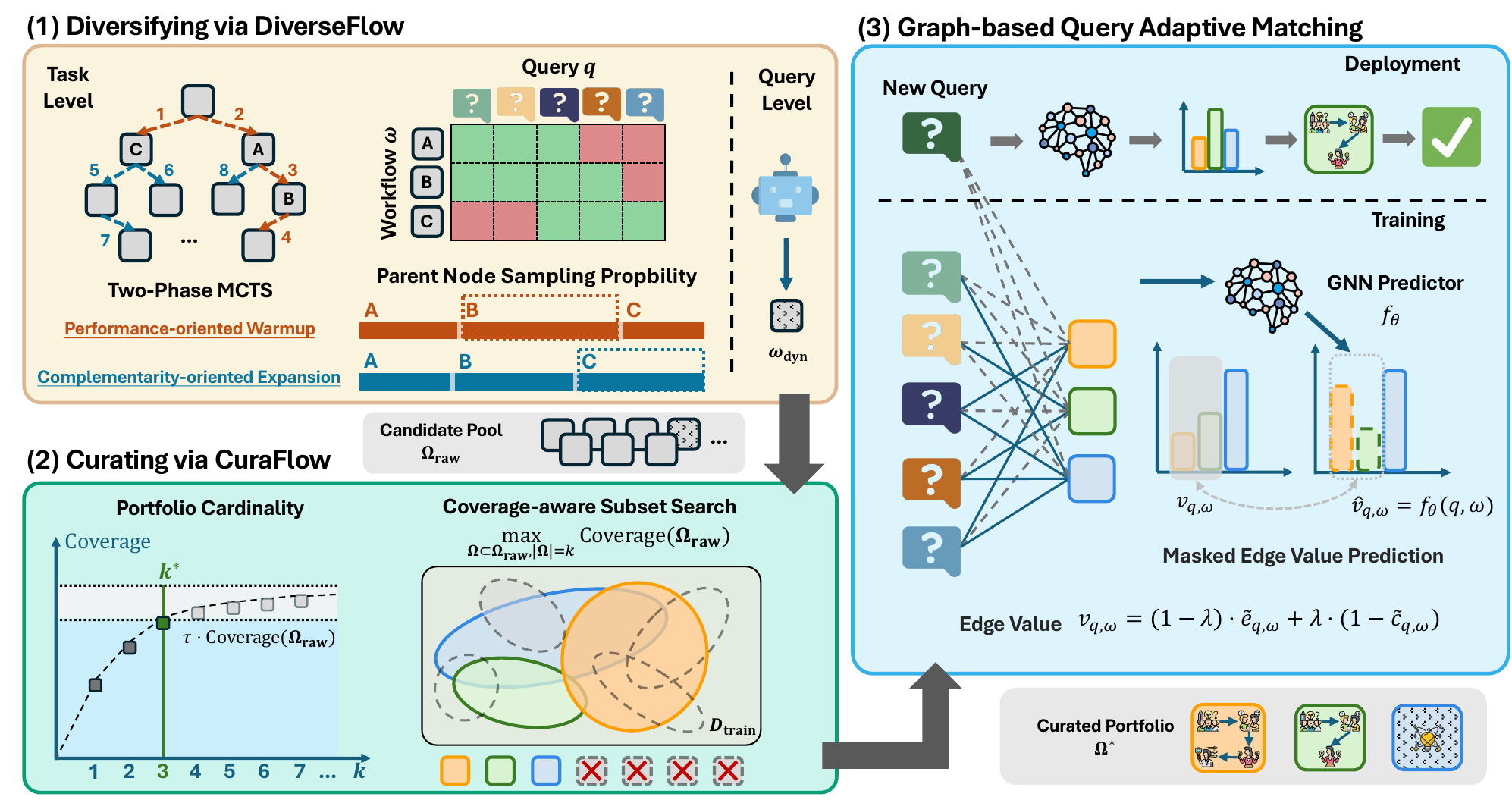} 
    \caption{Overview of \ours. DiverseFlow first builds a diverse raw pool $\Omega_{\text{raw}}$ through performance-oriented warm-up followed by coverage-oriented expansion. CuraFlow then selects a compact portfolio $\Omega^*$ that retains most attainable coverage while pruning redundant workflows. Given a new query, a bipartite query--workflow matcher predicts each portfolio member's utility under the performance--cost trade-off and executes the highest-scoring workflow.}
    \vspace{-8pt}
    \label{fig:pipeline} 
\end{figure}
\vspace{-8pt}
\enlargethispage{\baselineskip}

\paragraph{Performance-oriented Warm-up.}
During the first $N_0$ rounds, DiverseFlow runs a warm-up phase to initialize the candidate pool with high-performing individual workflows. At each round, a parent workflow is sampled from the current pool according to
\begin{equation}
\label{eq:aflow_sample}
    P(\omega_i)= \rho \cdot \frac{1}{n} + (1 - \rho) \cdot \frac{\exp\!\bigl(\alpha (\eta(\omega_i) - \eta_{\max})\bigr)}{\sum_{j=1}^{n} \exp\!\bigl(\alpha (\eta(\omega_j) - \eta_{\max})\bigr)},
\end{equation}
where $\eta_{\max} = \max_{\omega \in \Omega_{t-1}} \eta(\omega)$. We set the workflow sampling weight to $\eta(\omega)=\sum_{q\in D_{\text{train}}} e(\omega, q)$, so that workflows with stronger performance are more likely to be sampled as parents for further expansion. The value of $\alpha$ and $\rho$ control the exploration--exploitation trade-off. This warm-up phase initializes the pool with strong individual workflows. However, optimizing for individual performance alone is not sufficient for maximizing pool coverage as illustrated in Section~\ref{subsec:motivation_analysis}. High-performing workflows can still be redundant, while workflows that contribute unique coverage may be overlooked if their aggregate performance is less prominent.

\paragraph{Complementarity-oriented Expansion.}
To address this challenge, we switch DiverseFlow into a complementarity-oriented expansion phase after the warm-up stage.
The search structure remains unchanged.
For workflow weights $\eta(\omega)$ in \Cref{eq:aflow_sample}, we now use query-weighted performance $\sum_{q\in D_{\text{train}}}\overline{\mu}(q)\cdot e(\omega, q)$, where the query weight is given by
\begin{equation}
\label{eq:weighted_query_score}
    \overline{\mu}(q)=\frac{\mu(q)}{\sum_{q\in D_{\text{train}}} \mu(q)},\ \mu(q) = \frac{1}{1+\sum_{i=0}^{n} e(\omega_{i}, q)}.
\end{equation}
Here $\mu(q)$ reflects the difficulty of $q$ to the pool.
It reaches the maximum of $1$ when no workflow solves $q$.
Thus, the workflow with higher performance on "difficult" queries with greater $\mu(q)$ are more likely to be sampled as the parent to expand (a toy example provided in \Cref{fig:pipeline}), which steers the MCTS to search workflows that could solve uncovered questions. 

\paragraph{Finalizing Candidate Pool.}
After completing the two-phase search of DiverseFlow, we further augment the candidate pool by incorporating a query-level workflow generation method, ScoreFlow. We evaluate it by executing the generated workflow on its corresponding query and aggregating the resulting performance over $D_{\text{train}}$. We denote the final candidate pool by $\Omega_{\text{raw}}$, which includes both the workflows discovered by DiverseFlow and the additional ScoreFlow-based dynamic workflow. 

\subsection{Curating via CuraFlow}

\label{subsec:curation}

\paragraph{From a High-Coverage Pool to a Compact Portfolio.}
The diversified candidate pool produced in Stage~1 provides broad query coverage, but using the full pool directly is neither necessary nor desirable. Many workflows are partially redundant, and enlarging the pool yields diminishing marginal gains in coverage while increasing the cost of downstream matching and deployment complexity. We therefore introduce CuraFlow, a coverage-aware combinatorial curation procedure that consolidates the raw pool into a compact portfolio by searching for the smallest subset that preserves most of the attainable coverage.

\paragraph{Why DiverseFlow and CuraFlow are Complementary.}
Adding workflows in Stage~1  and removing workflows in Stage~2 optimize different objectives at different levels.
DiverseFlow is a candidate-generation stage: it deliberately expands $\Omega_{\text{raw}}$ so the system can discover a wide range of behaviors, including rare but useful workflows whose value may only become apparent when compared jointly with others. 
CuraFlow is a subset-selection stage: after this diversity has been exposed, it solves a combinatorial selection problem to keep the smallest portfolio that preserves the coverage achieved by the larger pool. 
In this sense, Stage~1 increases the \emph{recall} of useful workflow behaviors, while Stage~2 improves the \emph{precision} and deployability of the final portfolio. 
Without DiverseFlow, CuraFlow would have little complementary material to preserve; without CuraFlow, DiverseFlow would leave the system with an unnecessarily redundant and expensive pool. 
We provide a more detailed analysis in Appendix \ref{subsec:comparison_aflow_diverseflow_curaflow}.

\paragraph{Coverage-aware Portfolio Search.}
We initialize the portfolio cardinality $k$ at $1$ and expand it incrementally.
For each cardinality $k$, we identify the size-$k$ subset that maximizes coverage:
\begin{equation}
\Omega_k^* = \mathop{\arg\max}_{\Omega \subseteq \Omega_{\mathrm{raw}},\, |\Omega| = k} \mathrm{Coverage}(\Omega).
\label{eq:combinatorial_max}
\end{equation}
Exhaustive enumeration over all combinatorial subsets remains feasible when $k$ is modest and the candidate pool is of moderate size.
Among size-$k$ subsets with equal coverage, we prefer the one with the lowest mean pairwise correlation between members' per-query performance vectors. This encourages behavioral heterogeneity within the curated portfolio.

\paragraph{Optimal Cardinality for the Portfolio.}
To keep portfolio search tractable and control the scale of the matching stage, we use $k^*$ as the target size of the curated portfolio.
Because coverage is a submodular set function, its marginal gain diminishes as more workflows are added. As a result, a subset of size $k$ can retain most of the candidate pool's coverage while removing redundant workflows, even when $k$ is much smaller than $K$.
Practically, we choose the optimal portfolio size as
\begin{equation}
    k^*=\min\{k: \max_{\Omega\subset\Omega_{\text{raw}},|\Omega|=k}\text{Coverage}(\Omega)\geq\tau\cdot\text{Coverage}(\Omega_{\text{raw}})\}
\end{equation}
i.e., the minimum cardinality of a subset whose coverage exceeds a saturation ratio $\tau\in(0, 1]$ of the full candidate pool.
If $\tau$ falls on the plateau of the coverage curve for a given benchmark, the resulting $k^*$ may be slightly overestimated. In that case, we pass both $\Omega_k^*$ and $\Omega_{k-1}^*$ to the matching stage as alternative curated portfolios and treat the portfolio size as a tunable hyperparameter.

\subsection{Graph-Based Query-Adaptive Matching}
\label{subsec:query_matching}

\paragraph{Casting Workflow Selection as Edge-Value Prediction on a Bipartite Graph.}
Given the curated portfolio $\Omega^*$ obtained in Section~\ref{subsec:curation}, the last challenge is query-level assignment: for a new query, which portfolio member should execute it? A flat multiclass classifier over workflows would treat portfolio members as unrelated labels and force each query to have a unique winner, even though multiple workflows may be equally good under the deployment trade-off. We instead cast query-adaptive workflow selection as edge-value prediction on a query-workflow bipartite graph. This formulation makes the supervision pairwise by design, so the selector can exploit relational patterns across similar queries and similar workflows while naturally accommodating ties in workflow utility.

\paragraph{Bipartite Graph Construction and Node Representation Space.}
The bipartite graph is defined as $G = (D \cup {\Omega^*}, E)$, where $D$ is the set of query nodes and ${\Omega^*}$ is the set of workflow nodes. Each query node is initialized with an embedding of the query text, and each workflow node is initialized with an embedding of its workflow description (refer to Appendix~\ref{subsec:app_detail_of_ours} and \ref{sec:app_case_study} for more details). A heterogeneous GNN then maps the two node types into a shared hidden representation space through type-specific transformations, so that query semantics, workflow behavior, and their cross-type interactions can all inform the matching decision.
For every pair $(q,\omega) \in D_{\mathrm{train}} \times \Omega^*$, we create an edge $(q, \omega) \in E$ whose supervision value combines performance and cost:
\begin{equation}
v_{q,\omega} = (1 - \lambda) \cdot \tilde{e}_{q,\omega} 
            + \lambda \cdot (1 - \tilde{c}_{q,\omega}).
\label{eq:cost_aware_label}
\end{equation}
where $\tilde{e}_{q,\omega}$ and $\tilde{c}_{q,\omega}$ are the normalized performance and cost of workflow $\omega$ on query $q$ over $D_{\mathrm{train}}$, and $\lambda \in [0,1]$ controls the deployment trade-off between performance and token cost. 

\paragraph{Masked Training of a Lightweight Edge-Value Predictor.}
We train a heterogeneous $2$-layer GNN encoder followed by an MLP decoder to predict $\hat{v}_{q,\omega} = f_\theta(q,\omega)$. During training, we randomly sample minibatches of edges from $D_{\mathrm{train}} \times \Omega^*$ and mask their supervision values. The model receives the node representations together with the unmasked portion of the graph, propagates information over the bipartite structure, and predicts the masked edge values. We optimize a per-edge binary cross-entropy loss with soft targets on the masked edges. Because supervision is defined independently for every edge rather than as a single-winner class label for each query, ties between multiple workflows with the same $v_{q,\omega}$ are handled naturally: all tied edges simply share the same target value, and the model is not forced to arbitrarily separate them during training. 

\paragraph{Inductive Selecting for a New Query.} 
At deployment, a new query $q$ is embedded, added to the graph as a new query node, and connected to every workflow node in ${\Omega^*}$.
A single forward pass yields predicted edge values $f_\theta(q, \omega)$ for all $\omega \in \Omega^*$.
The selector is then defined as
$
\pi_\theta(q) = \arg\max_{\omega \in \Omega^*} f_\theta(q,\omega),
$
and the selected workflow $\pi_\theta(q)$ is executed on $q$ to produce the final answer. If multiple workflows receive the same predicted value, we break ties by preferring the lower-cost workflow on the training data, and use a fixed deterministic order only if the cost is also tied.
The inference cost of matching a new query is one GNN forward pass over a small graph, negligible compared to the cost of executing any workflow in the portfolio.
\section{Experiments}
\label{sec:experiments}

\definecolor{lightblue}{rgb}{0.9, 0.93, 0.99}     
\definecolor{lightbrown}{rgb}{0.99, 0.96, 0.93} 

\begin{table}[t]
\centering

\caption{Performance \& Cost Comparison with 13 baselines across 5 benchmarks. 
Performance results are bolded/underlined based on the ranking across all methods, while Cost metrics (unit: $\times 10^{-3}$ \$ per 1K tokens) are highlighted within the automated optimization block and \ours.}
\label{tab:performance_comparison}
\vspace{4pt}

{%
\renewcommand{\arraystretch}{1.3}
\setlength{\tabcolsep}{6pt}
\small
\resizebox{\textwidth}{!}{
\begin{tabular}{l cccccc cccccc}
\toprule[2.5pt]
\multirow{2.5}{*}{\textbf{Method}} & \multicolumn{2}{c}{\textbf{MATH}} & \multicolumn{2}{c}{\textbf{AMC}} & \multicolumn{2}{c}{\textbf{MBPP}} & \multicolumn{2}{c}{\textbf{DROP}} & \multicolumn{2}{c}{\textbf{MMLU Pro}} & \multicolumn{2}{c}{\textbf{Average}} \\
\cmidrule(lr){2-3} \cmidrule(lr){4-5} \cmidrule(lr){6-7} \cmidrule(lr){8-9} \cmidrule(lr){10-11} \cmidrule(lr){12-13}
& Perf. & Cost & Perf. & Cost & Perf. & Cost & Perf. & Cost & Perf. & Cost & Perf. & Cost \\
\midrule[1.5pt]

IO & 56.58 & 0.51 & 53.59 & 0.53 & 72.92 & 0.08 & 75.49 & 0.08 & 50.89 & 0.03 & 60.44 & 0.22 \\
\midrule[1.5pt]

\addlinespace[5pt] 
\rowcolor{lightblue}
\multicolumn{13}{c}{\textbf{\textit{Manually Designed Agentic Workflows}}} \\
\addlinespace[5pt] 

CoT & 55.76 & 0.53 & 52.47 & 0.54 & 74.29 & 0.08 & 77.96 & 0.11 & 51.35 & 0.04 & 60.98 & 0.23 \\
ComplexCoT & 54.39 & 0.62 & 49.52 & 0.64 & 71.68 & 0.10 & 77.74 & 0.33 & 64.03 & 0.35 & 63.85 & 0.42 \\
Self-Consistency & 57.82 & 3.38 & 53.89 & 3.44 & 73.98 & 0.52 & 78.33 & 0.74 & 50.93 & 0.34 & 61.49 & 1.51 \\
Multi-agent Debate & 58.92 & 7.66 & 53.82 & 7.96 & 74.68 & 0.86 & 78.65 & 1.64 & 57.56 & 0.88 & 63.86 & 3.45 \\
Self-Refine & 55.14 & 1.08 & 52.88 & 1.06 & 71.07 & 0.27 & 76.71 & 0.34 & 43.75 & 0.46 & 57.96 & 0.62 \\
MedPrompt & 56.69 & 5.64 & 55.19 & 5.49 & 71.85 & 0.61 & 81.77 & 1.75 & 52.79 & 0.63 & 62.77 & 2.58 \\
MultiPersona & 57.41 & 0.56 & 53.74 & 0.58 & 70.38 & 0.28 & 76.13 & 0.29 & 61.70 & 0.36 & 63.87 & 0.41 \\
\midrule[1.5pt]

\addlinespace[5pt]
\rowcolor{lightblue}
\multicolumn{13}{c}{\textbf{\textit{Automated Agentic Workflow Optimization Methods}}} \\
\addlinespace[5pt]

GPTSwarm & 54.94 & \underline{1.69} & 55.42 & 9.04 & 71.26 & \textbf{0.25} & 78.37 & \textbf{0.51} & 58.37 & 1.14 & 63.43 & 2.54 \\
ADAS & 56.17 & 2.95 & 47.33 & 3.22 & 72.34 & 0.77 & 81.17 & 0.98 & 59.71 & 1.05 & 63.22 & 1.71 \\
AgentSquare & 62.76 & \textbf{1.35} & 58.78 & \textbf{1.79} & 72.73 & \underline{0.52} & 76.84 & \underline{0.76} & 63.75 & \textbf{0.75} & 66.70 & \textbf{1.02} \\
AFlow (Qwen3-8B) & 63.24 & 3.23 & 62.75 & 2.51 & 79.37 & 1.02 & 77.08 & 1.05 & 64.62 & 1.47 & 68.56 & 1.79 \\
AFlow (GPT-4o) & 63.99 & 2.77 & \underline{63.77} & 4.86 & \underline{83.77} & 0.57 & 80.26 & 1.02 & \underline{65.61} & \underline{0.89} & \underline{70.40} & 1.95 \\
MaAS & \underline{65.02} & 2.43 & 63.36 & 2.70 & 79.08 & 2.04 & 76.64 & 1.64 & 62.31 & 1.29 & 68.09 & 1.90 \\
ScoreFlow & 62.00 & 2.36 & 60.15 & 3.51 & 83.63 & 1.70 & \underline{82.10} & 1.42 & 64.58 & 2.62 & 69.51 & 2.37 \\

\addlinespace[5pt]
\rowcolor{lightbrown}
\textbf{\ours} & \textbf{69.34} & 1.78 & \textbf{67.94} & \underline{2.49} & \textbf{84.26} & 1.62 & \textbf{83.49} & 1.06 & \textbf{67.40} & 1.52 & \textbf{73.40} & \underline{1.65} \\
\addlinespace[5pt]

\bottomrule[2.5pt]
\end{tabular}
}
}
\vspace{-8pt}

\end{table}

\subsection{Experimental Setup}
\label{subsec:experimental_setup}
\paragraph{Tasks and Benchmarks.}
We evaluate \ours on five public benchmarks covering four domains: \textbf{(1) math reasoning}, MATH \citep{MATH}, AMC \citep{Easy2hard}; \textbf{(2) code generation}, MBPP \citep{MBPP}; \textbf{(3) question answering}, MMLU Pro \citep{MMLU_Pro} and \textbf{(4) reading comprehension}, DROP \citep{DROP}. Following prior work, each benchmark is split into training and test sets with a $1{:}4$ ratio for workflow optimization and final evaluation. More details can be found in Appendix \ref{subsec:app_datasets}.

\paragraph{Baselines.} We compare \ours with two series of baselines: \textbf{(1)manually designed workflows}: Chain-of-Thought \citep{CoT}, ComplexCoT \citep{ComplexCoT}, Self-Consistency \citep{SelfConsistency}, Multi-agent Debate \citep{MultiAgentDebate}, Self-Refine \citep{SelfRefine}, MedPrompt \citep{MedPrompt}, MultiPersona \citep{MultiPersona}; \textbf{(2) automated workflow optimization methods}: GPTSwarm \citep{GPTSwarm}, ADAS \citep{ADAS}, AgentSquare \citep{AgentSquare}, AFlow \citep{AFlow}, MaAS \citep{MaAS}, ScoreFlow \citep{ScoreFlow}. 

\paragraph{LLM Backbones.}
For ADAS and AgentSquare, we use GPT-4o \citep{GPT-4o} as the optimizer LLM. For AFlow, we evaluate two variants using GPT-4o and Qwen3-8B \citep{Qwen3} as the optimizer LLM respectively, treating each as a separate baseline. For ScoreFlow, we use Qwen3-8B as the generator to design a workflow based on the given query. For \ours, we use Qwen3-8B as the optimizer at Stage 1.
To ensure a fair comparison, we use GPT-4o mini \citep{GPT-4omini} with a fixed temperature of 0 as the executor LLM and across all agentic workflows. More details can be found in Appendix \ref{subsec:app_model}.

\paragraph{Implementation Details.}
We run AFlow and ADAS for 30 search iteration rounds, and AgentSquare 
for 20. For ScoreFlow, we follow the official setting in their codebase and train its generator via Score-DPO for 3 epochs on each benchmark's training split
For \ours, Stage~1 runs the optimizer for 30 rounds with 7 warm-up rounds to construct a diverse candidate workflow pool. Stage~2 fixes the saturation ratio at $\tau = 0.96$ and curates two compact portfolios of sizes $k^*$ and $k^*-1$. Stage~3 then trains the query-adaptive selector over a grid of cost weights $\lambda$ on each dataset.
We provide more implementation details in Appendix \ref{subsec:app_baseline} and \ref{subsec:app_detail_of_ours}.

\subsection{Performance \& Cost Analysis}

\paragraph{Main Results.} Table~\ref{tab:performance_comparison} shows that \ours achieves the best average performance ($73.40$) among all methods and consistently ranks first across all five benchmarks. Compared with the strongest automated baseline, AFlow (GPT-4o) at $70.40$, \ours improves by $3.00$ absolute points ($4.26\%$ relative), despite using Qwen3-8B rather than GPT-4o as the optimizer. The gain over the strongest manually designed workflow, MultiPersona at $63.87$, is even larger at $9.53$ points ($14.92\%$ relative). At the benchmark level, \ours attains the best results on MATH ($69.34$), AMC ($67.94$), MMLU Pro ($67.40$), DROP ($83.49$), and MBPP ($84.26$), surpassing the next best method by clear margins on several reasoning-intensive tasks. Overall, these results show that the performance gains come from \ours's 3-stage framework itself, rather than simply from using a stronger optimizer model. See Appendix \ref{sec:app_case_study} for more detailed case study.

\begin{wrapfigure}{r}{0.55\textwidth}
    \vspace{-1.0\baselineskip}
    \centering
    \includegraphics[width=0.53\textwidth]{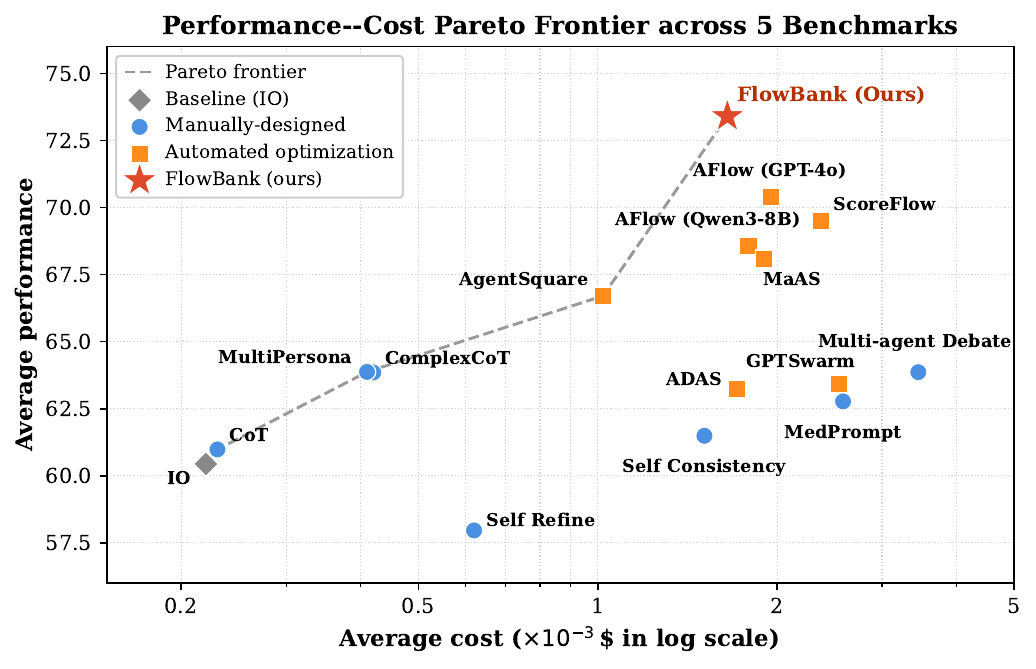}
    \vspace{-8pt}
    \captionsetup{hypcap=false}
    \caption{Performance--cost trade-off across all methods. \ours is on the Pareto frontier.}
    \label{fig:pareto_front}
    \vspace{-0.4\baselineskip}
\end{wrapfigure}
\paragraph{Performance--Cost Trade-off.} Beyond raw accuracy, \ours remains efficient. Its average inference cost is $1.65$, below AFlow (GPT-4o) at $1.95$ and ScoreFlow at $2.37$, while its average performance is higher than both. AgentSquare is cheaper at $1.02$, but it trails \ours by $6.70$ points in average performance, leaving it off the preferred utility frontier. As Figure~\ref{fig:pareto_front} shows, no compared method offers both higher average performance and lower average cost than \ours. This result shows that \ours improves both task effectiveness and the overall performance--cost balance. 

\subsection{Framework Analysis}

\paragraph{Ablation Study.}
Table~\ref{tab:ablation_3stage} compares the full system against four variants to isolate the contributions of each component. Compared with Row 2, the full system replaces AFlow with DiverseFlow in Stage~1, yielding a portfolio with higher oracle performance and, in turn, better final performance, which confirms the value of constructing a more diverse candidate pool. Compared with Row 3, CuraFlow improves over the top-$k$ accuracy set by using combinatorial search to directly maximize coverage, leading to a stronger oracle portfolio and higher downstream performance as well. Compared with Row 4, training the selector on the full candidate pool without curation causes noticeable degradation, suggesting that redundancy makes it harder for the selector to exploit the distinct strengths of individual workflows. 
Finally, compared with Row 5, the bipartite-graph based GNN selector consistently outperforms an independent flat MLP classifier, validating the benefit of explicitly modeling query--workflow interactions over a graph. 
Overall, the best results are achieved only when all three stages are activated together.

\begin{table}[!t]
    \centering
    \vspace{-4pt}
    \centering

\vspace{-7pt}
\caption{Ablation study of \ours on MATH and DROP. ``Oracle'' denotes the performance of an oracle selector on test data.}
\label{tab:ablation_3stage}
\vspace{6pt}

{%
\renewcommand{\arraystretch}{0.6}
\setlength{\tabcolsep}{9pt}
\small
\resizebox{0.85\textwidth}{!}{%
\begin{tabular}{ccccccc}
\toprule[1.5pt]
\multicolumn{3}{c}{\textbf{Method}} & \multicolumn{2}{c}{\textbf{MATH}} & \multicolumn{2}{c}{\textbf{DROP}} \\
\cmidrule(lr){1-3} \cmidrule(lr){4-5} \cmidrule(lr){6-7}
Stage 1 & Stage 2 & Stage 3 & Perf. & Oracle & Perf. & Oracle \\
\midrule[1pt]
$\checkmark$ & $\checkmark$ & $\checkmark$ & \textbf{69.34} & 83.74 & \textbf{83.49} & 90.47 \\
\texttimes\,(AFlow) & $\checkmark$ & $\checkmark$ & 67.69 & 79.01 & 82.98 & 89.95 \\
$\checkmark$ & \texttimes\,(Top-$k$ Accuracy set) & $\checkmark$ & 67.90 & 80.66 & 82.37 & 86.73 \\
$\checkmark$ & \texttimes\,(Full candidate pool) & $\checkmark$ & 68.11 & \textbf{92.04} & 82.45 & \textbf{96.59} \\
$\checkmark$ & $\checkmark$ & \texttimes\,(MLP Classifier) & 68.72 & 83.74 & 82.83 & 90.47 \\
\bottomrule[1.5pt]
\end{tabular}%
}
}
\vspace{-8pt}
\end{table}

\begin{wrapfigure}{r}{0.54\textwidth}
    \vspace{-0.5\baselineskip}
    \centering
    \includegraphics[width=0.52\textwidth]{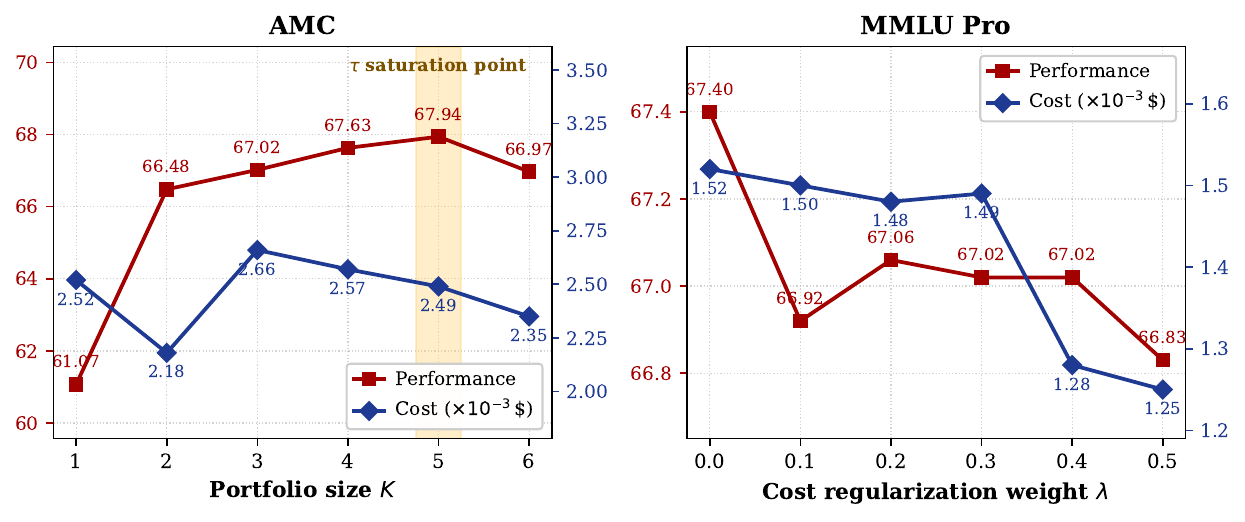}
    \vspace{-6pt}
    \captionsetup{hypcap=false}
    \caption{\textbf{Left:} Impact of portfolio size $k$ on AMC; \textbf{Right:} Impact of cost regularization weight $\lambda$ on MMLU Pro.}
    \label{fig:hp_combined}
    \vspace{-0.5\baselineskip}
\end{wrapfigure}
\paragraph{Hyperparameter Effects.} Figure~\ref{fig:hp_combined} studies two key hyperparameters in \ours and shows that AMC performance increases from $61.07$ at $K{=}1$ to $67.94$ at $K{=}5$ before dipping slightly at $K{=}6$. This pattern supports the importance of choosing the saturation point for a compact, diverse portfolio for the third stage to learn the selector over a portfolio with an appropriate number of workflows. On MMLU Pro, larger $\lambda$ steadily lowers inference cost while performance stays stable for $\lambda \in [0.1, 0.4]$ and drops at $\lambda{=}0.5$, showing the balance between effect and token efficiency. Overall, \ours is robust to moderate changes in $\lambda$ and maintains a strong performance--cost trade-off without heavy tuning.
\section{Related Work}
\paragraph{LLM-based Agentic Workflow.}
Recent work has shown that LLMs can benefit substantially from explicit workflow design rather than a single prompt-response pass. Representative manually designed strategies include Chain-of-Thought, Self-Consistency, debate-based collaboration, and iterative refinement, as well as more specialized designs such as MedPrompt, MultiPersona and MetaGPT~\citep{CoT, SelfConsistency, MultiAgentDebate, SelfRefine, MedPrompt, MultiPersona, MetaGPT}. These methods demonstrate the value of workflow structure and prompting, but they usually rely on manually designed workflows.

\paragraph{Automated Agentic Workflow Optimization.}
Another line of work studies how to automatically discover and optimize workflows, including task-level offline workflow search methods such as GPTSwarm, ADAS, AgentSquare, AFlow and Weak-for-strong, as well as more adaptive query-level online generation systems such as MAS-GPT, ScoreFlow, MaAS, MasRouter and DyFlow~\citep{GPTSwarm, ADAS, AgentSquare, AFlow, MAS-GPT, ScoreFlow, MaAS, MasRouter, Weak-for-strong, DyFlow}. These methods reduce manual effort and often improve average task performance, but most either optimize for a single final workflow or shift substantial workflow construction to inference time. In contrast, \ours constructs a compact portfolio of complementary workflows offline and learns to select the best one for each query at inference time, making our method closer to a precompute-and-reuse framework than to either single-workflow optimization or fully online workflow generation.

\section{Conclusion}

We revisit agentic workflow optimization from a different perspective. Rather than choosing between a single workflow for all queries and synthesizing a new workflow for every query, we ask how to recover query-level adaptivity through offline computation and reusable workflow assets. \ours operationalizes this idea with three stages---\emph{Diversifying}, \emph{Curating}, and \emph{Matching}---that discover complementary workflows, distill them into a compact portfolio, and assign each query to the right workflow at inference time. Across five benchmarks, this precompute-and-reuse strategy delivers the strongest average performance among the evaluated methods while maintaining competitive cost. More detailed discussion about limitations and future directions can be found at Appendix \ref{sec:app_add_discussion}.
\section{Acknowledgment}

Yuan, Deng, Chakraborty and Huang are supported by DARPA HR001124S0029-AIQ-FP-019,  National Science Foundation TRAILS Institute (2229885). Private support was provided by Open Philanthropy and Apple. The Authors acknowledge the National Artificial Intelligence Research Resource (NAIRR) Pilot  for contributing to this research result.
\clearpage
\bibliography{main}

\appendix
\section{Notations}
This appendix summarizes the main symbols used in Section~\ref{sec:motivation} and Section~\ref{sec:method}.

\begin{table}[H]
\centering
\small
\caption{Notation summary for the formulation and the three stages of \ours.}
\label{tab:notation}
\begin{tabular}{>{\centering\arraybackslash}p{0.20\textwidth}>{\centering\arraybackslash}p{0.72\textwidth}}
\toprule
\textbf{Symbol} & \textbf{Meaning} \\
\midrule
$D$ & Task dataset. \\
$D_{\text{train}}$ & Training split. \\
$q$ & Input query. \\
$\omega$ & Agentic workflow. \\
$\omega(q)$ & Workflow prediction on $q$. \\
$e(\omega, q)$ & Workflow performance on $q$. \\
$c(\omega, q)$ & Workflow token cost on $q$. \\
$\omega_{\text{stat}}$ & Static workflow shared by all queries. \\
$G_{\phi}$ & Meta-generator for dynamic workflows. \\
$\omega^{q}_{\text{dyn}} = G_{\phi}(q)$ & Dynamic workflow generated for $q$. \\
$\Omega$ & A set consists of workflows. \\
$\Omega_{\text{raw}}$ & Raw candidate workflow pool. \\
$\mu(q)$ & Complementarity-aware weight of query $q$. \\
$\mathrm{Coverage}(\Omega)$ & Set-level utility of subset $\Omega$. \\
$k^*$ & Chosen portfolio size. \\
$\tau$ & Coverage saturation ratio. \\
$\Omega^*$ & Final curated portfolio. \\
$G=(D \cup \Omega^*, E)$ & Query-workflow bipartite graph. \\
$v_{q,\omega}$ & Edge supervision value for $(q,\omega)$. \\
$\tilde e_{q,\omega}$ & Normalized effect of $\omega$ on $q$. \\
$\tilde c_{q,\omega}$ & Normalized cost of $\omega$ on $q$. \\
$\lambda$ & Effect-cost trade-off weight. \\
$f_{\theta}(q,\omega)$ & Predicted value of matching $q$ with $\omega$. \\
$\pi_{\theta}(q)$ & Selected workflow for query $q$. \\
\bottomrule
\end{tabular}
\end{table}
\section{Algorithm Table}

Algorithm~\ref{alg:diverseflow} summarizes Stage~1, where DiverseFlow maintains a growing workflow pool and shifts from performance-oriented search to coverage-oriented search. Algorithm~\ref{alg:curation} summarizes CuraFlow, the combinatorial curation procedure in Stage~2. Algorithm~\ref{alg:matching} describes Stage~3, where a lightweight graph-based selector predicts the best workflow for each query.

\begin{algorithm}[htbp]
\caption{DiverseFlow for Building a High-Coverage Candidate Pool}
\label{alg:diverseflow}
\small
\begin{algorithmic}
\Require Training split $D_{\text{train}}$, search budget $N$, warm-up rounds $N_0$
\Ensure Raw workflow pool $\Omega_{\text{raw}}$
\State Initialize workflow pool $\Omega_0$ with seed workflows
\For{$t=1$ to $N$}
    \If{$t \leq N_0$}
        \State Score each workflow in $\Omega_{t-1}$ by its aggregate performance on $D_{\text{train}}$
    \Else
        \State Compute query weights $\mu(q)$ from the current pool coverage on $D_{\text{train}}$
        \State Re-score each workflow in $\Omega_{t-1}$ by weighted performance under $\mu(q)$
    \EndIf
    \State Sample a parent workflow $\omega^{\text{parent}}$ from $\Omega_{t-1}$
    \State Ask the optimizer LLM to propose a new workflow $\omega^{\text{new}}$ conditioned on $\omega^{\text{parent}}$ and search experience
    \State Execute $\omega^{\text{new}}$ on $D_{\text{train}}$ to obtain per-query effect and token cost
    \State Compare $\omega^{\text{new}}$ against $\omega^{\text{parent}}$ under the current scoring rule and record success/failure feedback
    \State Update the workflow pool: $\Omega_t \leftarrow \Omega_{t-1} \cup \{\omega^{\text{new}}\}$
\EndFor
\State \Return $\Omega_{\text{raw}} \leftarrow \Omega_N$
\end{algorithmic}
\end{algorithm}

\begin{algorithm}[htbp]
\caption{CuraFlow for Combinatorial Portfolio Curation}
\label{alg:curation}
\small
\begin{algorithmic}
\Require Raw candidate pool $\Omega_{\text{raw}}$, training split $D_{\text{train}}$, saturation ratio $\tau$
\Ensure Candidate curated portfolios for the matching stage
\For{$k=1$ to $|\Omega_{\text{raw}}|$}
    \State Enumerate size-$k$ subsets $\Omega \subseteq \Omega_{\text{raw}}$
    \State Compute $\mathrm{Coverage}(\Omega)$ on $D_{\text{train}}$
    \State Select $\Omega_k^* = \arg\max_{|\Omega|=k} \mathrm{Coverage}(\Omega)$
    \State Break ties by preferring lower mean pairwise correlation between per-query effect vectors
\EndFor
\State Find the smallest $k^*$ such that
\Statex \hspace{1.5em}$\mathrm{Coverage}(\Omega_{k^*}^*) \geq \tau \cdot \mathrm{Coverage}(\Omega_{\text{raw}})$
\State Form candidate curated portfolios from $\Omega_{k^*}^*$ and, when applicable, $\Omega_{k^*-1}^*$
\State \Return candidate portfolios $\Omega^*$ for Stage~3
\end{algorithmic}
\end{algorithm}

\begin{algorithm}[t]
\caption{Query-Adaptive Matching over the Curated Portfolio}
\label{alg:matching}
\small
\begin{algorithmic}
\Require Curated portfolio $\Omega^*$, training split $D_{\text{train}}$, trade-off coefficient $\lambda$, new query $q$
\Ensure Selected workflow $\pi_{\theta}(q)$
\State Build bipartite graph $G=(D_{\text{train}} \cup \Omega^*, E)$ with complete query-workflow edges
\ForAll{$(q',\omega) \in D_{\text{train}} \times \Omega^*$}
    \State Compute normalized effect $\tilde e_{q',\omega}$ and normalized cost $\tilde c_{q',\omega}$
    \State Assign edge supervision value $v_{q',\omega} = (1-\lambda)\tilde e_{q',\omega} + \lambda(1-\tilde c_{q',\omega})$
\EndFor
\State Train a heterogeneous 2-layer GNN with an MLP decoder to predict $f_{\theta}(q',\omega)$ from node embeddings and observed edges
\State Add the new query $q$ as a query node and connect it to every workflow in $\Omega^*$
\State Perform one forward pass to obtain $f_{\theta}(q,\omega)$ for all $\omega \in \Omega^*$
\State Select the best workflow:
\Statex \hspace{1.5em}$\pi_{\theta}(q) \leftarrow \arg\max_{\omega \in \Omega^*} f_{\theta}(q,\omega)$
\State Break ties by preferring the lower-cost workflow
\State \Return $\pi_{\theta}(q)$
\end{algorithmic}
\end{algorithm}
\section{Detailed Experimental Setup}
\label{sec:app_detailed_experimental_set}

\subsection{Datasets and Splits}
\label{subsec:app_datasets}

We evaluate all methods on five public benchmarks spanning four domains: MATH and AMC for mathematical reasoning, MBPP for code generation, MMLU Pro for question answering, and DROP for reading comprehension. Following prior workflow-optimization work \citep{ADAS, AFlow, ScoreFlow, Weak-for-strong}, we split each benchmark into training and test sets with a $1{:}4$ ratio, where the training split is used for workflow search, curation, or selector training, and the test split is reserved for final evaluation. Table~\ref{tab:data_statics} summarizes the resulting dataset statistics.

\begin{table}[!t]
    \centering
\vspace{-6pt}
\caption{Statistics of the training and test splits for the five benchmarks.}
\label{tab:data_statics}
\vspace{4pt}

{%
\renewcommand{\arraystretch}{0.95}
\setlength{\tabcolsep}{12pt}
\small
\begin{tabular}{lcc}
\toprule[1.5pt]
\textbf{Dataset} & \textbf{Training} & \textbf{Testing} \\
\midrule[1pt]
MATH     & 119 & 486  \\
AMC      & 165 & 655  \\
MBPP     & 86  & 341  \\
DROP     & 200 & 800  \\
MMLU Pro & 260 & 1040 \\
\bottomrule[1.5pt]
\end{tabular}
}
\vspace{-6pt}
\end{table}

For MATH, we follow previous works \citep{ADAS, AFlow} in selecting 605 problems from four typical problem types (Combinatorics \& Probability, Number Theory, Pre-algebra, Pre-calculus) at difficulty level 5. For AMC, we use the Easy2Hard benchmark \citep{Easy2hard} and retain problems whose IRT difficulty ratings fall in $[0.25, 0.40]$, resulting in 841 problems in total. For MMLU Pro, we exclude the duplicated math category and then uniformly sample 100 problems from each of the remaining 13 categories, yielding 1{,}300 problems. 

\subsection{Model Configurations}
\label{subsec:app_model}

We use GPT-4o \citep{GPT-4o} and Qwen3-8B \citep{Qwen3} as optimizer-side large language models in different methods, depending on each baseline's design. Specifically, ADAS and AgentSquare use GPT-4o as the optimizer LLM. For AFlow, we report two variants, one with GPT-4o and the other with Qwen3-8B as the optimizer, and treat them as separate baselines. ScoreFlow uses Qwen3-8B as the generator for workflow design, and \ours also uses Qwen3-8B as the optimizer in Stage~1.

To keep execution-time comparison fair across all agentic workflows, we use GPT-4o mini \citep{GPT-4omini} with temperature fixed to 0 as the executor LLM throughout the experiments. For GPT-4o, we use the default configuration. For Qwen3-8B, we use temperature 0.7 in non-reasoning mode.

\subsection{Baselines and Setup}
\label{subsec:app_baseline}

We compare \ours against two categories of baselines. The first category consists of manually designed workflows, including Chain-of-Thought \citep{CoT}, ComplexCoT \citep{ComplexCoT}, Self-Consistency \citep{SelfConsistency}, Multi-agent Debate \citep{MultiAgentDebate}, Self Refine \citep{SelfRefine}, MedPrompt \citep{MedPrompt}, and MultiPersona \citep{MultiPersona}. The second category consists of automated workflow optimization methods, including four task-level methods: GPTSwarm \citep{GPTSwarm}, ADAS \citep{ADAS}, AgentSquare \citep{AgentSquare}, AFlow \citep{AFlow} and two query-level methods: MaAS \citep{MaAS} and ScoreFlow \citep{ScoreFlow}.

For manually designed workflows, we implement each method according to its standard formulation. For CoT, we prepend the instruction ``let's think step by step'' before answer generation. For ComplexCoT, we use a more structured prompt that instructs the model to identify variables, formulate a plan, execute the solution, and verify the result. For Self-Consistency, we sample $n=5$ CoT responses and aggregate them using an LLM-based ensemble vote. For Self-Refine, we first generate an initial CoT solution and then perform up to three rounds of review and revision, with early stopping when the reviewer judges the current answer to be correct. For Multi-Agent Debate, three agents independently produce CoT solutions, exchange their answers for two rounds of debate, and then produce a final prediction through ensemble voting. For MultiPersona, a single LLM call identifies task-relevant personas and simulates a multi-turn collaboration among them within one generation. For MedPrompt, we retrieve the top-3 most similar few-shot exemplars using embedding similarity, sample five conditioned solutions, and ensemble the resulting predictions.

For search-based baselines, we follow the standard optimization budgets and implementation settings whenever applicable. Specifically, we run AFlow and ADAS for 30 search iterations, and AgentSquare for 20 iterations. For GPTSwarm, we adopt the edge-optimization framework with REINFORCE training, following the official implementation settings, with a learning rate of 0.1, an initial connection probability of 0.5, and a per-problem moving-average baseline. For MaAS, we follow the default configuration reported in the original paper~\cite{MaAS}, using a four-layer query-aware controller trained for four sampling repetitions and GPT-4o mini as the TextGrad-based prompt-refinement model. For ScoreFlow, we follow the official setting in its released codebase and train the generator with Score-DPO for three epochs on the training split of each benchmark. All methods are evaluated under the same executor model and test protocol described above.

\subsection{Implementation Details of \ours}
\label{subsec:app_detail_of_ours}

Our implementation follows the three-stage framework introduced in Section~\ref{sec:method}. In Stage~1, we run DiverseFlow with Qwen3-8B as the optimizer for 30 rounds on the training split, using 7 warmup rounds before the main diversified search. After that, we also incorporate ScoreFlow into the final candidate pool $\Omega_{\text{raw}}$. Specifically, for each training query, ScoreFlow first generates a workflow and then executes it on that query. We record the resulting prediction and total token cost, including both workflow generation and execution, and treat these query-wise outputs as ScoreFlow's candidate performance. In this way, ScoreFlow can be represented in the same result--cost format as other workflows for downstream curation.

In Stage~2, we fix the saturation ratio to $\tau = 0.96$ and run coverage-aware curation on each benchmark to obtain two compact portfolio candidates with sizes $k^*$ and $k^*-1$, respectively. In practice, the raw candidate pool remains moderate, with $|\Omega_{\text{raw}}| = 30 + 1 = 31$, where the additional candidate comes from ScoreFlow. Moreover, as shown in Figure~\ref{fig:all_benchmarks_coverage_k_curve}, performance typically saturates once the portfolio size reaches at most $k=6$. Since both the candidate-pool size and the effective search range are small, exhaustive subset search is computationally affordable in our setting. This design gives a near-saturated portfolio together with a slightly smaller alternative, which supports the subsequent performance--cost trade-off analysis.

For larger candidate pools or target portfolio sizes, CuraFlow admits a scalable greedy variant in place of exhaustive subset enumeration. Starting from the empty set, this variant grows the portfolio one workflow at a time, and at each step adds the candidate with the largest marginal increase in Coverage. Under our definition of Coverage as the average query-wise best performance over a workflow set, the resulting objective is monotone submodular as long as per-query performance is nonnegative. Therefore, for any fixed budget $k$, the greedy portfolio achieves at least a $(1-1/e)$ fraction of the optimal size-$k$ Coverage. In practice, one can first build the nested greedy sequence and then select the smallest $k$ whose Coverage reaches the saturation threshold, preserving the core behavior of CuraFlow while scaling to substantially larger search spaces.

In Stage~3, following \cite{GraphRouter}, we first use GPT-4o to generate a natural-language description for each workflow in the curated portfolio (specific examples provided in Appendix \ref{sec:app_case_study}). We then encode both user queries and workflow descriptions using OpenAI's \texttt{text-embedding-3-small} model, construct the bipartite query--workflow graph, and train the GNN-based selector over a grid of cost regularization weights $\lambda \in [0, 0.1, 0.2, 0.3, 0.4, 0.5]$ on each benchmark. At inference time, the selector predicts the most suitable workflow for each query under the learned performance--cost trade-off. All GNN-based selectors are trained on a single NVIDIA RTX A5000 GPU with 24~GB of memory. Each run uses approximately 300--400~MB of GPU memory, allowing us to evaluate multiple hyperparameter configurations concurrently on a single device. A 500-step training run takes approximately 7--8 minutes, and a complete hyperparameter sweep can be finished within one hour of wall-clock time on a single GPU.
\section{Case Study}
\label{sec:app_case_study}

\subsection{AMC}

\begin{itemize}
  \item \textbf{AMC W1}

\begin{tcolorbox}[
    title=Description, 
    colback=gray!5, 
    colframe=colorAMC,
    breakable
]
Three-way diverse-prompt ensemble (no code verification). Custom operator generates 3 independent solutions under different instruction styles --- SOLVE\_PROMPT (general), ANALYTICAL\_SOLVE\_PROMPT (deductive analysis), and VERBOSE\_SOLVE\_PROMPT (verbose explanation) --- then ScEnsemble selects the most consistent answer via majority voting. Four LLM calls in total. Lighter than W2/W3 because it does not use Programmer-based verification. It is suited to AMC problems where reframing the question through different prompt styles surfaces the correct reasoning path more reliably than a single fixed prompt.
\end{tcolorbox}

\begin{tcblisting}{
  title=Python Code,
  colback=gray!5,
  colframe=colorAMC,
  enhanced jigsaw,
  breakable,
  listing only,
  listing engine=listings,
  listing options={
    language=Python,
    basicstyle=\ttfamily\scriptsize,
    keywordstyle=\color{blue},
    commentstyle=\color{forestgreen},
    stringstyle=\color{darkred},
    showstringspaces=false,
    breaklines=true,
  }
}
SOLVE_PROMPT = """
You are a math problem solver. Please solve the following problem step by step, showing all your work and providing a final answer in a box.

Problem:
{problem}

Solution:
"""

ANALYTICAL_SOLVE_PROMPT = """
You are a math problem solver. Please solve the following problem step by step, showing all your work and providing a final answer in a box. Focus on breaking down the problem into smaller parts and analyze each part carefully.

Problem:
{problem}

Solution:
"""

VERBOSE_SOLVE_PROMPT = """
You are a math problem solver. Please solve the following problem step by step, showing all your work and providing a final answer in a box. Be as detailed as possible in your explanation, and make sure to explain the reasoning behind each step.

Problem:
{problem}

Solution:
"""


class Workflow:
    def __init__(self, name, llm_config, dataset):
        self.name = name
        self.dataset = dataset
        self.llm = create_llm_instance(llm_config)
        self.custom = operator.Custom(self.llm)
        self.sc_ensemble = operator.ScEnsemble(self.llm)

    async def __call__(self, problem: str):
        # Three diverse-style candidates, then ensemble-vote
        prompts = [SOLVE_PROMPT, ANALYTICAL_SOLVE_PROMPT, VERBOSE_SOLVE_PROMPT]
        solutions = [await self.custom(input=problem, instruction=p) for p in prompts]
        best_solution = await self.sc_ensemble(solutions=solutions, problem=problem)
        return best_solution, self.llm.usage_tracker.get_summary()["total_cost"]
\end{tcblisting}

\begin{tcolorbox}[
    title=Assigned Query Example, 
    colback=gray!5, 
    colframe=colorAMC
]
Suppose $a$ and $b$ are single-digit positive integers chosen independently and at random. What is the probability that the point $(a,b)$ lies above the parabola $y=ax^2-bx$?
\end{tcolorbox}

  \item \textbf{AMC W2}

\begin{tcolorbox}[
    title=Description, 
    colback=gray!5, 
    colframe=colorAMC,
    breakable
]
Three-way analytical Custom ensemble with code verification. Generates 3 independent analytical solutions using a Custom operator with a focused analytical prompt, ensembles them via ScEnsemble majority voting, then verifies the best solution with a Programmer operator that executes Python code. Effective for AMC problems that benefit from diverse analytical reasoning plus computational validation.
\end{tcolorbox}

\begin{tcblisting}{
  title=Python Code,
  colback=gray!5,
  colframe=colorAMC,
  enhanced jigsaw,
  breakable,
  listing only,
  listing engine=listings,
  listing options={
    language=Python,
    basicstyle=\ttfamily\scriptsize,
    keywordstyle=\color{blue},
    commentstyle=\color{forestgreen},
    stringstyle=\color{darkred},
    showstringspaces=false,
    breaklines=true,
  }
}
ANALYTICAL_SOLVE_PROMPT = """
You are a math problem solver. Please solve the following problem step by step, showing all your work and providing a final answer in a box. Focus on breaking down the problem into smaller parts and analyze each part carefully. Make sure to explain your reasoning and avoid assumptions.

Problem:
{problem}

Solution:
"""


class Workflow:
    def __init__(self, name, llm_config, dataset):
        self.name = name
        self.dataset = dataset
        self.llm = create_llm_instance(llm_config)
        self.custom = operator.Custom(self.llm)
        self.sc_ensemble = operator.ScEnsemble(self.llm)
        self.programmer = operator.Programmer(self.llm)

    async def __call__(self, problem: str):
        # Three samples from the same analytical prompt, then ensemble-vote
        solutions = [
            await self.custom(input=problem, instruction=ANALYTICAL_SOLVE_PROMPT)
            for _ in range(3)
        ]
        best_solution = await self.sc_ensemble(solutions=solutions, problem=problem)

        # Verify the ensemble pick with executed Python code
        final_solution = await self.programmer(problem=problem, analysis=best_solution)
        return final_solution, self.llm.usage_tracker.get_summary()["total_cost"]
\end{tcblisting}

\begin{tcolorbox}[
    title=Assigned Query Example, 
    colback=gray!5, 
    colframe=colorAMC
]
Two distinct numbers a and b are chosen randomly from the set $\{2, 2^2, 2^3, ..., 2^{25}\}$. What is the probability that $\mathrm{log}_a b$ is an integer?
\end{tcolorbox}

  \item \textbf{AMC W3}

\begin{tcolorbox}[
    title=Description, 
    colback=gray!5, 
    colframe=colorAMC,
    breakable
]
Three-way diverse-prompting ensemble with code verification. Generates 3 solutions using different prompt strategies (logical reasoning, thorough explanation, concise structure), ensembles via ScEnsemble, then verifies the chosen solution with a Programmer operator. Effective for AMC problems where varied reasoning approaches increase coverage of the solution space and code execution catches arithmetic slips.
\end{tcolorbox}

\begin{tcblisting}{
  title=Python Code,
  colback=gray!5,
  colframe=colorAMC,
  enhanced jigsaw,
  breakable,
  listing only,
  listing engine=listings,
  listing options={
    language=Python,
    basicstyle=\ttfamily\scriptsize,
    keywordstyle=\color{blue},
    commentstyle=\color{forestgreen},
    stringstyle=\color{darkred},
    showstringspaces=false,
    breaklines=true,
  }
}
LOGICAL_PROMPT = "You are a math problem solver. Please solve the following problem step by step, showing all your work and providing a final answer in a box. Focus on logical reasoning and clarity."

THOROUGH_PROMPT = "You are a math problem solver. Please solve the following problem step by step, showing all your work and providing a final answer in a box. Be thorough and explain your thought process."

CONCISE_PROMPT = "You are a math problem solver. Please solve the following problem step by step, showing all your work and providing a final answer in a box. Make sure your solution is concise and well-structured."


class Workflow:
    def __init__(self, name, llm_config, dataset):
        self.name = name
        self.dataset = dataset
        self.llm = create_llm_instance(llm_config)
        self.custom = operator.Custom(self.llm)
        self.sc_ensemble = operator.ScEnsemble(self.llm)
        self.programmer = operator.Programmer(self.llm)

    async def __call__(self, problem: str):
        # Three diverse-style candidates, then ensemble-vote
        prompts = [LOGICAL_PROMPT, THOROUGH_PROMPT, CONCISE_PROMPT]
        solutions = [await self.custom(input=problem, instruction=p) for p in prompts]
        best_solution = await self.sc_ensemble(solutions=solutions, problem=problem)

        # Verify with executed Python code
        final_solution = await self.programmer(problem=problem, analysis=best_solution)
        return final_solution, self.llm.usage_tracker.get_summary()["total_cost"]
\end{tcblisting}

\begin{tcolorbox}[
    title=Assigned Query Example, 
    colback=gray!5, 
    colframe=colorAMC
]
Kiana has two older twin brothers.  The product of their three ages is 128.  What is the sum of their three ages?
\end{tcolorbox}

  \item \textbf{AMC W4}

\begin{tcolorbox}[
    title=Description, 
    colback=gray!5, 
    colframe=colorAMC,
    breakable
]
Classify-then-solve conditional routing workflow. First classifies the problem into a category (geometry, combinatorics, number theory, or algebra) via a Custom operator with a classifier prompt, then routes the query to a domain-specialist Custom call whose prompt is tailored for that category. Two LLM calls total --- the cheapest WF in the AMC pool. Effective when domain-specific reasoning strategies significantly outperform generic prompting; weak when the classifier mis-routes.
\end{tcolorbox}

\begin{tcblisting}{
  title=Python Code,
  colback=gray!5,
  colframe=colorAMC,
  enhanced jigsaw,
  breakable,
  listing only,
  listing engine=listings,
  listing options={
    language=Python,
    basicstyle=\ttfamily\scriptsize,
    keywordstyle=\color{blue},
    commentstyle=\color{forestgreen},
    stringstyle=\color{darkred},
    showstringspaces=false,
    breaklines=true,
  }
}
CLASSIFY_PROMPT = """Analyze this math problem and classify it into exactly one category. Output ONLY the category name, nothing else.

Categories:
- ALGEBRA: equations, inequalities, functions, polynomials, sequences
- GEOMETRY: shapes, angles, areas, volumes, coordinate geometry
- COMBINATORICS: counting, probability, permutations, combinations
- NUMBER_THEORY: divisibility, primes, modular arithmetic, digits
"""

ALGEBRA_PROMPT = """You are an algebra specialist. Solve this problem using algebraic techniques:
- Set up equations from the problem conditions
- Solve systematically (substitution, elimination, factoring)
- Check your answer satisfies all original conditions

Put your final answer in \\boxed{{}}.
"""

GEOMETRY_PROMPT = """You are a geometry specialist. Solve this problem using geometric reasoning:
- Draw a mental picture and label key points/lengths/angles
- Apply relevant theorems (Pythagorean, similar triangles, circle properties)
- Use coordinate geometry if the problem involves positions

Put your final answer in \\boxed{{}}.
"""

COMBINATORICS_PROMPT = """You are a combinatorics specialist. Solve this problem using counting techniques:
- Identify what is being counted or the probability space
- Choose the right technique: multiplication principle, combinations, inclusion-exclusion, complementary counting
- Double-check by considering small cases if possible

Put your final answer in \\boxed{{}}.
"""

NUMBER_THEORY_PROMPT = """You are a number theory specialist. Solve this problem using number-theoretic techniques:
- Consider divisibility rules, prime factorizations, modular arithmetic
- Look for patterns in small cases
- Apply relevant theorems (CRT, Fermat's little theorem, etc.)

Put your final answer in \\boxed{{}}.
"""


class Workflow:
    """Classify problem type, then route to a domain-specialist prompt."""
    def __init__(self, name, llm_config, dataset):
        self.name = name
        self.dataset = dataset
        self.llm = create_llm_instance(llm_config)
        self.custom = operator.Custom(self.llm)

    async def __call__(self, problem: str):
        # Classify into one of four math domains
        category = (await self.custom(input=problem, instruction=CLASSIFY_PROMPT)).strip().upper()

        # Route to a specialist prompt (algebra is the default)
        if "GEOMETRY" in category:
            specialist = GEOMETRY_PROMPT
        elif "COMBINATORICS" in category:
            specialist = COMBINATORICS_PROMPT
        elif "NUMBER" in category:
            specialist = NUMBER_THEORY_PROMPT
        else:
            specialist = ALGEBRA_PROMPT

        solution = await self.custom(input=problem, instruction=specialist)
        return solution, self.llm.usage_tracker.get_summary()["total_cost"]
\end{tcblisting}

\begin{tcolorbox}[
    title=Assigned Query Example, 
    colback=gray!5, 
    colframe=colorAMC
]
What is the volume of a cube whose surface area is twice that of a cube with volume 1?
\end{tcolorbox}

  \item \textbf{AMC W5}
\begin{tcolorbox}[
    title=Description, 
    colback=gray!5, 
    colframe=colorAMC,
    breakable
]
ScoreFlow query-level workflow generator. Unlike fixed-graph workflows that apply one static computation graph to every query, ScoreFlow synthesizes a custom workflow on-the-fly for each individual query at inference time. An optimizer LLM inspects the query and produces a tailored graph before execution. The exposed cost includes both the generated workflow's runtime cost and ScoreFlow's own generation overhead, making it more expensive than fixed-graph candidates. Strong overall accuracy, especially on queries where fixed routing heuristics fail; provides genuine per-query adaptability that cannot be replicated by any single fixed graph.
\end{tcolorbox}

\begin{tcblisting}{
  title=Python Code,
  colback=gray!5,
  colframe=colorAMC,
  enhanced jigsaw,
  breakable,
  listing only,
  listing engine=listings,
  listing options={
    language=Python,
    basicstyle=\ttfamily\scriptsize,
    keywordstyle=\color{blue},
    commentstyle=\color{forestgreen},
    stringstyle=\color{darkred},
    showstringspaces=false,
    breaklines=true,
  }
}
class Workflow:
    def __init__(
        self,
        config,
        problem
    ) -> None:
        self.problem = problem
        self.config = create(config)
        self.gid = None
        self.custom = operator.Custom(self.config, self.gid, self.problem)
        self.sc_ensemble = operator.ScEnsemble(self.config, self.gid, self.problem)
        self.programmer = operator.Programmer(self.config, self.gid, self.problem)
        self.review = operator.Review(self.config, self.gid, self.problem)
        
    async def run_workflow(self):
        """
        This is a workflow graph.
        """
        step1 = await self.custom(instruction="Can you break down the problem into smaller steps and explain the reasoning behind each step?")
        step2 = await self.custom(instruction="Can you solve the problem by providing a detailed solution?")
        solution = await self.sc_ensemble(solutions=[step1, step2])
        final_solution = await self.programmer(analysis=solution)
        reviewed_solution = await self.review(pre_solution=final_solution)
        return reviewed_solution
\end{tcblisting}

\begin{tcolorbox}[
    title=Assigned Query Example, 
    colback=gray!5, 
    colframe=colorAMC
]
A singles tournament had six players. Each player played every other player only once, with no ties. If Helen won 4 games, Ines won 3 games, Janet won 2 games, Kendra won 2 games and Lara won 2 games, how many games did Monica win?
\end{tcolorbox}

\end{itemize}

\subsection{MBPP}

\begin{itemize}
  \item \textbf{MBPP W1}

\begin{tcolorbox}[
    title=Description, 
    colback=gray!5, 
    colframe=colorMBPP,
    breakable
]
ScoreFlow query-level workflow generator. Unlike fixed-graph workflows that apply one static computation graph to every query, ScoreFlow synthesizes a custom workflow on-the-fly for each individual query at inference time. An optimizer LLM inspects the query and produces a tailored graph before execution. The exposed cost includes both the generated workflow's runtime cost and ScoreFlow's own generation overhead, making it more expensive than fixed-graph candidates. Strong overall accuracy, especially on queries where fixed routing heuristics fail; provides genuine per-query adaptability that cannot be replicated by any single fixed graph.
\end{tcolorbox}

\begin{tcblisting}{
  title=Python Code,
  colback=gray!5,
  colframe=colorMBPP,
  enhanced jigsaw,
  breakable,
  listing only,
  listing engine=listings,
  listing options={
    language=Python,
    basicstyle=\ttfamily\scriptsize,
    keywordstyle=\color{blue},
    commentstyle=\color{forestgreen},
    stringstyle=\color{darkred},
    showstringspaces=false,
    breaklines=true,
  }
}
class Workflow:
    def __init__(
        self,
        config,
        problem
    ) -> None:
        self.problem = problem
        self.config = create(config)
        self.code_generate = operator.CustomCodeGenerate(self.config, self.problem)
        self.sc_ensemble = operator.ScEnsemble(self.config, self.problem)
        self.test = operator.Test(self.config, self.problem)

    async def run_workflow(self):
        """
        This is a workflow graph.
        """
        solution_1 = await self.code_generate(instruction="Can you analyze this problem step by step and generate a function to find the nth decagonal number?")
        solution_2 = await self.code_generate(instruction="Can you think of another way to generate a function to find the nth decagonal number?")
        solution_3 = await self.code_generate(instruction="Can you provide a different approach to generate a function to find the nth decagonal number?")
        
        ensembled_solution = await self.sc_ensemble(solutions=[solution_1, solution_2, solution_3])
        
        tested_solution = await self.test(solution=ensembled_solution)
        
        return tested_solution
\end{tcblisting}

\begin{tcolorbox}[
    title=Assigned Query Example, 
    colback=gray!5, 
    colframe=colorMBPP
]
Write a function to find the nth decagonal number.

def is\_num\_decagonal(n):
\end{tcolorbox}

  \item \textbf{MBPP W2}

\begin{tcolorbox}[
    title=Description, 
    colback=gray!5, 
    colframe=colorMBPP,
    breakable
]
Clarify-Solve-Iterate pipeline without ensembling. First a Custom operator rewrites the problem through a clarify prompt to pin down any ambiguous requirements. Then a single CustomCodeGenerate call under a step-by-step reasoning prompt produces one candidate function implementation. The Test operator runs the problem's unit checks; if they pass, return immediately. On failure the workflow enters a bounded refine loop: up to three Custom refine passes each consume the last failed solution plus the error trace and emit a corrected version, with Test run after each iteration. No majority voting, no parallel candidates --- commits to one reasoning path and iterates on errors via focused refinement. Up to 5 LLM calls in the worst case, typically 2 on problems the first-pass solution already solves. Architecturally a direct contrast to W1's per-query-generated workflow: this is a fixed graph that's been tuned to one particular problem-solving strategy.
\end{tcolorbox}

\begin{tcblisting}{
  title=Python Code,
  colback=gray!5,
  colframe=colorMBPP,
  enhanced jigsaw,
  breakable,
  listing only,
  listing engine=listings,
  listing options={
    language=Python,
    basicstyle=\ttfamily\scriptsize,
    keywordstyle=\color{blue},
    commentstyle=\color{forestgreen},
    stringstyle=\color{darkred},
    showstringspaces=false,
    breaklines=true,
  }
}
CLARIFY_PROMPT = "Refine the problem understanding and clarify the requirements."

SOLVE_PROMPT = "Solve the problem step by step and provide a clear solution."

REFINE_PROMPT = "Refine the solution to fix any errors or improve clarity."


class Workflow:
    def __init__(self, name, llm_config, dataset):
        self.name = name
        self.dataset = dataset
        self.llm = create_llm_instance(llm_config)
        self.custom = operator.Custom(self.llm)
        self.custom_code_generate = operator.CustomCodeGenerate(self.llm)
        self.test = operator.Test(self.llm, dataset=self.dataset)

    async def __call__(self, problem: str, entry_point: str):
        # Clarify the problem, then generate an initial solution
        refined_problem = await self.custom(input=problem, instruction=CLARIFY_PROMPT)
        solution = await self.custom_code_generate(
            problem=refined_problem, entry_point=entry_point, instruction=SOLVE_PROMPT
        )
        test_result = await self.test(problem=problem, solution=solution, entry_point=entry_point)

        # Up to 3 refinement passes; stop early as soon as tests pass
        for _ in range(3):
            if test_result['result']:
                break
            refined_solution = await self.custom(
                input=test_result['solution'], instruction=REFINE_PROMPT
            )
            test_result = await self.test(
                problem=problem, solution=refined_solution, entry_point=entry_point
            )

        return test_result['solution'], self.llm.usage_tracker.get_summary()["total_cost"]
\end{tcblisting}

\begin{tcolorbox}[
    title=Assigned Query Example, 
    colback=gray!5, 
    colframe=colorMBPP
]
Write a function to extract all the adjacent coordinates of the given coordinate tuple.

def get\_coordinates(test\_tup):
\end{tcolorbox}

\end{itemize}

\subsection{DROP}

\begin{itemize}
  \item \textbf{DROP W1}

\begin{tcolorbox}[
    title=Description, 
    colback=gray!5, 
    colframe=colorDROP,
    breakable
]
Conditional routing workflow based on question-type classification. First classifies the question into NUMERIC, COMPARE, DATE, SPAN, or other. For NUMERIC/COMPARE questions, routes to QANumerical --- a DROP-tailored Programmer variant that generates Python code to extract numbers and compute the answer in a sandboxed subprocess. For DATE questions, uses a Custom operator with a date-extraction prompt. For SPAN questions, uses a Custom operator with a span-extraction prompt. All other questions fall back to AnswerGenerate. This selective routing applies QANumerical's strength (code-based computation) only where it fits, avoiding it on text-span or date questions where it would overfit to numeric patterns. Low cost (2 LLM calls: classify + solve).
\end{tcolorbox}

\begin{tcblisting}{
  title=Python Code,
  colback=gray!5,
  colframe=colorDROP,
  enhanced jigsaw,
  breakable,
  listing only,
  listing engine=listings,
  listing options={
    language=Python,
    basicstyle=\ttfamily\scriptsize,
    keywordstyle=\color{blue},
    commentstyle=\color{forestgreen},
    stringstyle=\color{darkred},
    showstringspaces=false,
    breaklines=true,
  }
}
CLASSIFY_PROMPT = """Classify the question into ONE of these categories. Output ONLY the category name (a single word).

Categories:
- NUMERIC --- asks for a count, sum, difference, percentage, year, or any calculation.
- SPAN --- asks "who", "what", "which" to extract a text span (name, place, entity) from the passage.
- DATE --- asks for a specific date.
- COMPARE --- asks to compare two things (e.g., "which is greater", "longer", "earlier").

Output format: just one word, e.g. NUMERIC
"""

SPAN_PROMPT = """Read the passage and find the exact text span that answers the question.
Output ONLY the text span, no explanation, no formatting.
"""

DATE_PROMPT = """Read the passage and find the exact date that answers the question.
Output ONLY the date in the format used by the passage, no explanation.
"""


class Workflow:
    """Classify question type, then route to the best-suited operator."""
    def __init__(self, name, llm_config, dataset):
        self.name = name
        self.dataset = dataset
        self.llm = create_llm_instance(llm_config)
        self.custom = operator.Custom(self.llm)
        self.answer_generate = operator.AnswerGenerate(self.llm)
        self.qa_numerical = operator.QANumerical(self.llm)

    async def __call__(self, problem: str):
        cls = (await self.custom(input=problem, instruction=CLASSIFY_PROMPT)).strip().upper()

        if "NUMERIC" in cls or "COMPARE" in cls:
            solution = await self.qa_numerical(problem=problem, analysis="None")
        elif "DATE" in cls:
            solution = await self.custom(input=problem, instruction=DATE_PROMPT)
        elif "SPAN" in cls:
            solution = await self.custom(input=problem, instruction=SPAN_PROMPT)
        else:
            solution = await self.answer_generate(input=problem)

        return solution, self.llm.usage_tracker.get_summary()["total_cost"]
\end{tcblisting}

\begin{tcolorbox}[
    title=Assigned Query Example, 
    colback=gray!5, 
    colframe=colorDROP
]
Passage: The median age in the city was 22.1 years. 10.1\% of residents were under the age of 18; 56.2\% were between the ages of 18 and 24; 16.1\% were from 25 to 44; 10.5\% were from 45 to 64; and 7\% were 65 years of age or older. The gender makeup of the city was 64.3\% male and 35.7\% female.
Question: How many in percent weren't between the ages of 18 and 24?
Answer:
\end{tcolorbox}

  \item \textbf{DROP W2}

\begin{tcolorbox}[
    title=Description, 
    colback=gray!5, 
    colframe=colorDROP,
    breakable
]
ScoreFlow query-level workflow generator. Unlike fixed-graph workflows that apply one static computation graph to every query, ScoreFlow synthesizes a custom workflow on-the-fly for each individual query at inference time. An optimizer LLM inspects the query and produces a tailored graph before execution. The exposed cost includes both the generated workflow's runtime cost and ScoreFlow's own generation overhead, making it more expensive than fixed-graph candidates. Strong overall accuracy, especially on queries where fixed routing heuristics fail; provides genuine per-query adaptability that cannot be replicated by any single fixed graph.
\end{tcolorbox}

\begin{tcblisting}{
  title=Python Code,
  colback=gray!5,
  colframe=colorDROP,
  enhanced jigsaw,
  breakable,
  listing only,
  listing engine=listings,
  listing options={
    language=Python,
    basicstyle=\ttfamily\scriptsize,
    keywordstyle=\color{blue},
    commentstyle=\color{forestgreen},
    stringstyle=\color{darkred},
    showstringspaces=false,
    breaklines=true,
  }
}
class Workflow:
    def __init__(
        self,
        config,
        problem
    ) -> None:
        self.problem = problem
        self.agent = create(config)
        self.custom = operator.Custom(self.agent, self.problem)
        self.answer_generate = operator.AnswerGenerate(self.agent, self.problem)
        self.sc_ensemble = operator.ScEnsemble(self.agent, self.problem)
        self.review = operator.Review(self.agent, self.problem)

    async def run_workflow(self):
        """
        This is a workflow graph.
        """
        initial_solution = await self.answer_generate()
        reviewed_solution = await self.review(pre_solution=initial_solution)
        ensembled_solution = await self.sc_ensemble(solutions=[initial_solution, reviewed_solution])
        final_solution = await self.review(pre_solution=ensembled_solution)
        
        return final_solution
\end{tcblisting}

\begin{tcolorbox}[
    title=Assigned Query Example, 
    colback=gray!5, 
    colframe=colorDROP
]
Passage: The median age in the city was 22.1 years. 10.1\% of residents were under the age of 18; 56.2\% were between the ages of 18 and 24; 16.1\% were from 25 to 44; 10.5\% were from 45 to 64; and 7\% were 65 years of age or older. The gender makeup of the city was 64.3\% male and 35.7\% female.
Question: Which age group had the second most people?
Answer:
\end{tcolorbox}

\end{itemize}

\section{Supplementary Experiment Results}
\label{sec:app_supp_experiment}

\begin{figure}[!htbp]
    \centering
    \includegraphics[width=\linewidth]{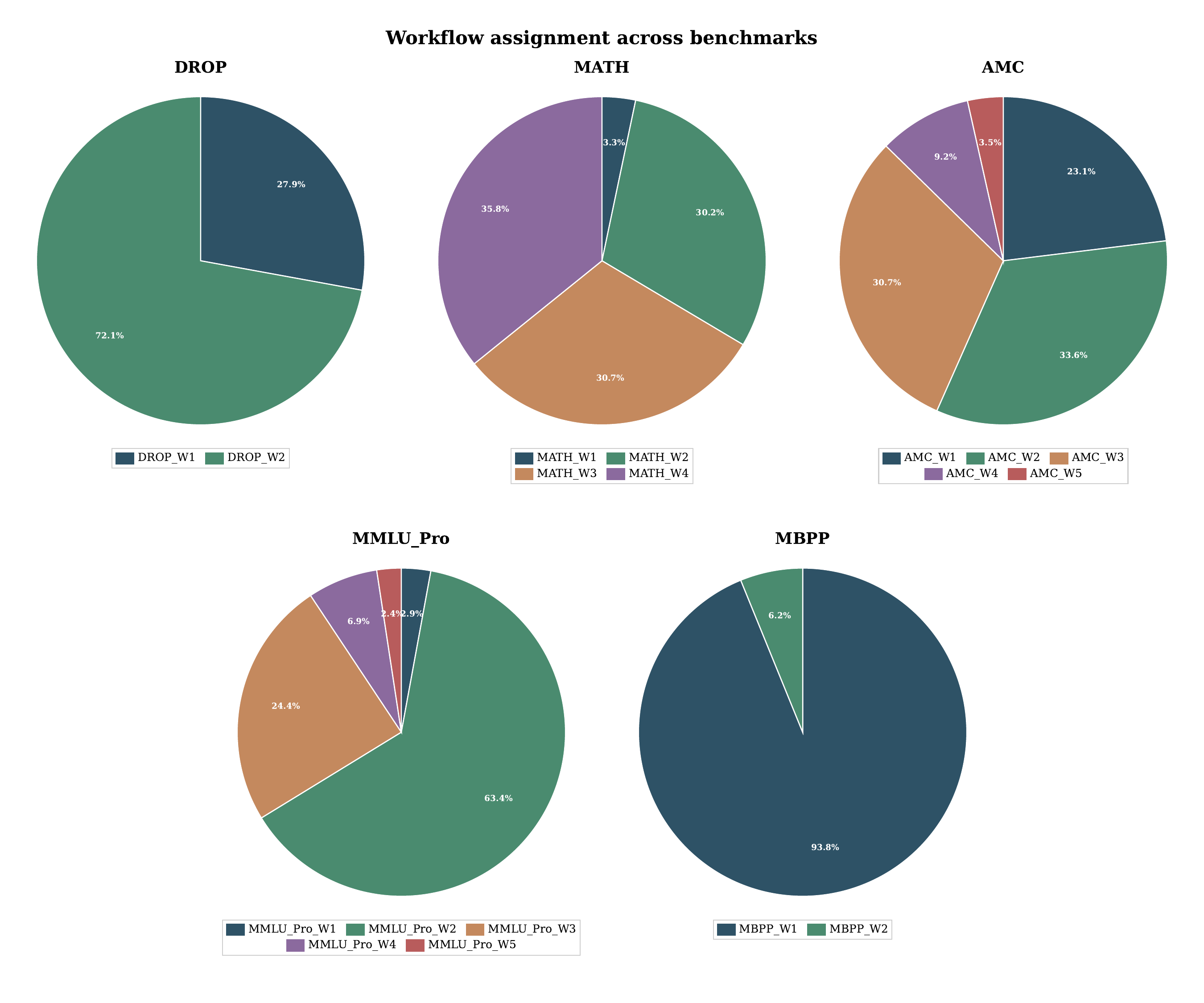}
    \caption{Workflow assignment distribution of \ours across datasets.}
    \label{fig:combined_routing_distribution}
\end{figure}

\subsection{Distribution of Workflow Assignment on Each Dataset}

As shown in Figure \ref{fig:combined_routing_distribution}, \ours successfully assigns different queries to different members in the curated portfolio over five benchmarks, demonstrating \ours's query adaptivity.

\subsection{Coverage-\texorpdfstring{$k$}{k} Curves in CuraFlow on five benchmarks}
\begin{figure}[!htbp]
    \centering
    \includegraphics[width=0.9\linewidth]{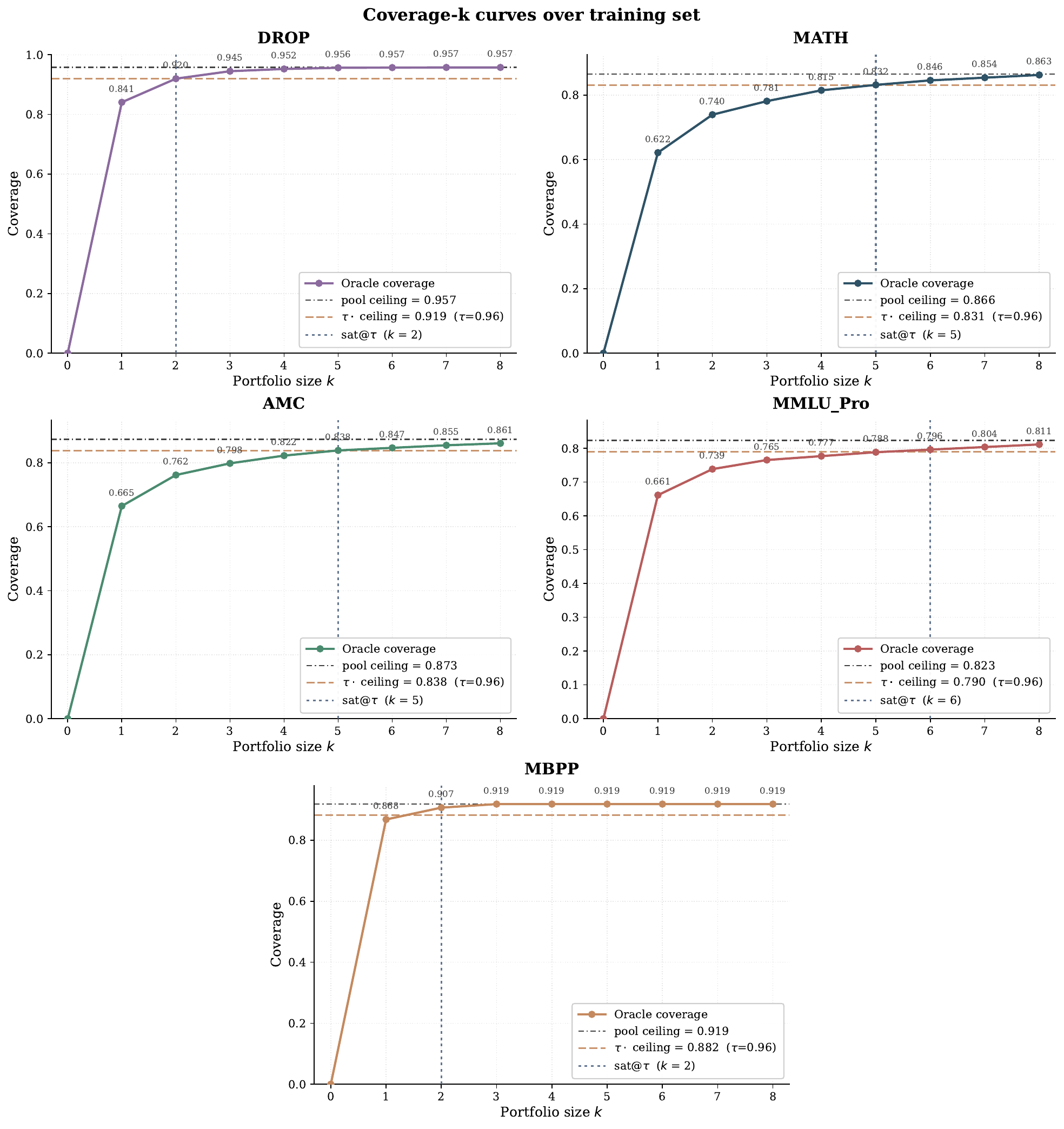}
    \caption{Coverage-$k$ curves of CuraFlow across five benchmarks.}
    \label{fig:all_benchmarks_coverage_k_curve}
\end{figure}

Figure~\ref{fig:all_benchmarks_coverage_k_curve} summarizes the Coverage-$k$ curves obtained by CuraFlow across all five benchmarks. We observe a consistent pattern: coverage improves rapidly when a small number of workflows are added, and then gradually saturates as $k$ increases. This indicates that much of the attainable set-level utility can be captured by a compact portfolio, which supports our design choice of curating a small workflow bank instead of keeping the entire candidate pool. We also observe noticeable variation across datasets: some benchmarks saturate quickly, while others continue to benefit from adding more workflows, suggesting that the degree of workflow complementarity is task dependent. 

\subsection{Comparison of Coverage-\texorpdfstring{$k$}{k} Curves for AFlow, DiverseFlow and CuraFlow}
\label{subsec:comparison_aflow_diverseflow_curaflow}
\begin{figure}[!htbp]
    \centering
    \includegraphics[width=0.8\linewidth]{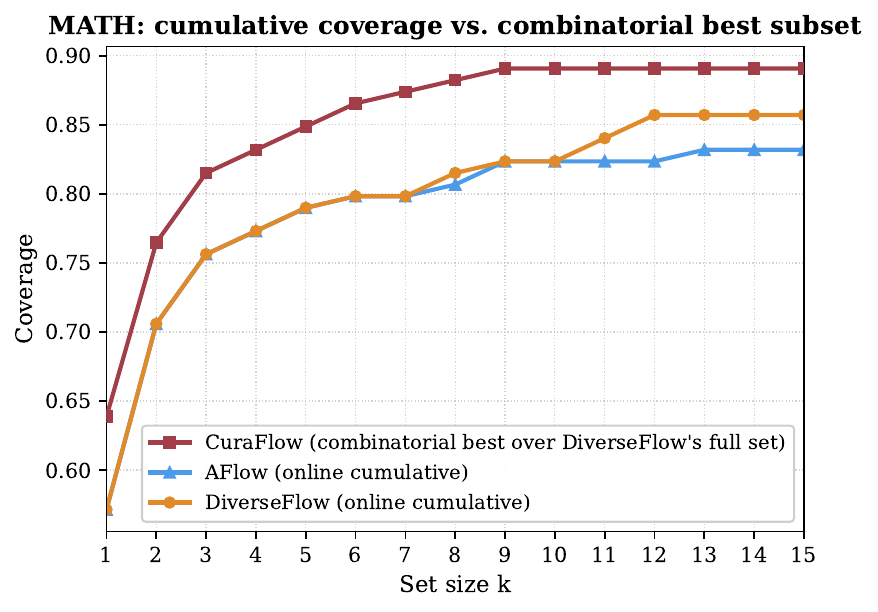}
    \caption{Controlled comparison of Coverage-\texorpdfstring{$k$}{k} curves for AFlow and DiverseFlow (online cumulative), and CuraFlow (combinatorial best over DiverseFlow's full candidate set) on MATH.}
    \label{fig:aflow_diverseflow_curaflow_coverage_k_curve}
\end{figure}

To better understand why DiverseFlow and CuraFlow are complementary, we conduct a controlled analysis on MATH that directly echoes the ``Why DiverseFlow and CuraFlow are Complementary'' discussion in Section~\ref{subsec:curation}. We run AFlow and DiverseFlow for the same 20 optimization rounds, while constraining the first seven rounds of DiverseFlow to produce exactly the same workflows as AFlow. We then compare their online cumulative set coverage during optimization (the blue and orange curves in Figure~\ref{fig:aflow_diverseflow_curaflow_coverage_k_curve}). After DiverseFlow finishes all 20 rounds, we further apply CuraFlow to the resulting workflow pool generated from DiverseFlow and perform post-hoc max-coverage search; this curve is shown as the red line in Figure~\ref{fig:aflow_diverseflow_curaflow_coverage_k_curve}.

Figure~\ref{fig:aflow_diverseflow_curaflow_coverage_k_curve} shows two key findings. First, after the shared first seven rounds, DiverseFlow consistently achieves higher online coverage than AFlow, indicating that diversity-aware exploration is indeed more effective at discovering complementary workflow sets. Second, although online exploration increases set size, it cannot anticipate which later workflows will best complement earlier ones, so stopping early still yields suboptimal sets. CuraFlow addresses this limitation by searching for the max-coverage subset over DiverseFlow's fully explored candidate pool, allowing a much smaller set to match or even exceed the coverage of the online cumulative set at the same size~$k$. This explains why the two stages are not contradictory: DiverseFlow discovers complementary candidates, and CuraFlow distills them into a compact high-coverage portfolio.

\section{Additional Discussion}
\label{sec:app_add_discussion}

\subsection{Limitations}
\label{subsec:limitation}
One main limitation of \ours is that the curated portfolio passed to Stage~3 is fixed once constructed. 
While this design improves deployment efficiency and simplifies inference-time routing, it also means that workflows discovered after curation are not automatically incorporated into the bank. Extending \ours into a continually expanding flow bank that grows as new workflows become available is a natural direction we leave to future work.
In addition, the saturation ratio $\tau$ and the cost trade-off $\lambda$ are lightweight hyperparameters that may benefit from per-benchmark tuning. Although Section~\ref{sec:experiments} shows that \ours is robust within a moderate range, fully automating their selection remains an open question.

\subsection{Broader Impact}
\label{subsec:impact}
\ours can have several positive practical impacts. By reusing a compact portfolio of high-utility workflows instead of generating or executing fully dynamic multi-agent systems for every query, it can substantially reduce token consumption and thus lower monetary and environmental costs in large-scale deployments of multi-agent systems. In addition, the framework is naturally expandable: the workflow bank can continually grow by incorporating workflows discovered from different optimization methods, datasets, domains, or external workflow sources, which may make it a flexible infrastructure layer for future agentic systems. At the same time, \ours may also introduce negative impacts. If the bank is populated with workflows collected from heterogeneous sources, it may inherit hidden biases, unsafe reasoning patterns, or licensing and attribution issues from those sources, and these problems may be amplified when such workflows are reused at scale.

\subsection{Future Work}

More broadly, our results suggest that workflows should be treated less as one-off artifacts and more as reusable memory. A natural next step is to move from a fixed portfolio to a continually expanding flow bank that can recognize when existing workflows are insufficient, incorporate newly discovered workflows, and improve routing under distribution shift. We hope this perspective helps shift the field from searching for one universally best workflow toward building reusable workflow libraries that grow in coverage, adaptability, and efficiency over time. One promising direction is to connect FlowBank with recent memory-based agent frameworks such as A-Mem~\citep{A-mem}, ArcMemo~\citep{ho2025arcmemo}, MemoryBank~\citep{Memorybank}, and ReasoningBank~\citep{reasoningbank}, so that it stores not only reusable workflows but also structured experience about when they succeed, fail, and should be adapted.
\section{Additional Related Works}

\paragraph{Static Agentic Workflow Optimization.}
Static agentic workflow optimization focuses on automatically improving agentic workflows and multi-agent architectures to reduce reliance on manually designed prompting and coordination strategies.
These approaches typically formulate agentic systems as structured computational graphs and optimize their prompts, topologies, communication patterns, or execution procedures through search, evolutionary algorithms, architecture optimization, or preference-based learning.
Early studies such as GPTSwarm~\citep{GPTSwarm} formulate language-agent systems as optimizable graphs and improve multi-agent collaboration through graph-based optimization.
ADAS~\citep{ADAS} further investigates automated agentic system design by performing program-level search over workflow structures and operators.
Building upon these ideas, AFlow~\citep{AFlow} introduces an MCTS-based framework for automated workflow generation and demonstrates strong reasoning performance across multiple benchmarks.
Mass~\citep{Mass} studies the joint optimization of prompts and agent topologies, emphasizing the importance of structural coordination in multi-agent systems.
Several subsequent works further improve workflow optimization efficiency and diversity.
EvoFlow~\citep{EvoFlow} adopts evolutionary strategies to iteratively evolve diverse workflows during optimization, while AdaptFlow~\citep{AdaptFlow} introduces meta-learning techniques to improve workflow adaptation across tasks and environments.
Preference-based optimization methods have also attracted growing attention.
SPOGW~\citep{SPOGW} proposes a score-based preference optimization framework using group-wise workflow comparisons to improve workflow quality.
In addition, FLORA~\citep{FLORA} studies surrogate performance modeling by using graph neural networks to predict workflow effectiveness, enabling more efficient workflow evaluation and search.
Despite their effectiveness, most existing methods ultimately optimize toward a single final workflow or architecture for deployment.
In contrast, our work focuses on constructing a compact portfolio of complementary workflows and adaptively selecting among them at inference time, enabling a better trade-off between deployment efficiency and query-level adaptivity.

\paragraph{Query-Adaptive Workflow Generation.}
Query-adaptive workflow generation focuses on improving adaptability at inference time by dynamically generating or orchestrating workflows for queries that require different reasoning strategies, coordination structures, or agent interactions.
Instead of deploying a single static workflow for all inputs, these methods dynamically generate, select, or route workflows at inference time to improve task-specific adaptability and reasoning performance.
MAS-GPT~\citep{MAS-GPT} trains large language models to directly generate multi-agent systems conditioned on the input query, enabling query-specific workflow construction through meta-generation. 
MaAS~\citep{MaAS} formulates workflow discovery as an architecture search problem and leverages a supernet-based strategy to efficiently explore large workflow design spaces.
ScoreFlow~\citep{ScoreFlow} further introduces score-based preference optimization for training workflow generators, allowing the system to synthesize customized workflows tailored to individual queries. 
FlowSteer~\citep{FlowSteer} formulates interactive workflow orchestration as an end-to-end reinforcement learning problem, optimizing adaptive workflow control policies through interaction feedback.
While these methods improve query-level flexibility, they often incur substantial online computation cost due to per-query workflow generation or orchestration. In contrast, our work seeks to recover much of the benefit of query adaptivity through a precompute-and-reuse paradigm, where a compact portfolio of complementary workflows is optimized offline and adaptively selected at inference time.

\paragraph{Analysis of Multi-Agent Systems.}
Analysis of multi-agent systems focuses on examining the foundations, efficiency, and necessity of complex orchestration strategies, asking when multi-agent systems and query-adaptive workflows are truly beneficial. These works investigate issues such as uncertainty propagation, coordination inefficiency, redundant workflow complexity, and the trade-offs between single-agent, static-workflow, and dynamically generated multi-agent systems, providing important insights into the practical limitations of agentic workflows.
Zhao et al.~\citep{On-the-Uncertainty} study uncertainty propagation and reliability in LLM-based multi-agent systems, showing how interactions among agents can amplify instability and affect downstream reasoning quality.
Li~\citep{WhenandWhen} investigates the conditions under which a strong single agent augmented with specialized skills can match or outperform multi-agent systems, revealing that additional coordination is not always necessary. 
Similarly, Xu et al.~\citep{Rethinking} revisit the empirical value of multi-agent workflows and demonstrate that carefully designed single-agent baselines can remain highly competitive across many tasks.
Several works further analyze the efficiency and scalability of complex orchestration strategies. 
El et al.~\citep{Inefficiencies} study the computational inefficiencies introduced by meta-agent-based workflow design and question whether increasingly sophisticated orchestration mechanisms justify their additional overhead. 
Wang et al.~\citep{Do-we-always-need-query-level} systematically analyze the necessity of query-level workflow generation and show that static workflows can remain competitive in many scenarios despite the growing popularity of dynamic workflow synthesis. 
Luo et al.~\citep{AgentArk} further explore whether the collaborative intelligence emerging from multi-agent systems can be distilled into a single LLM agent, studying the relationship between distributed coordination and compact agent representations.
Collectively, these works suggest that increasing workflow complexity or online adaptivity does not necessarily translate into better efficiency or scalability. In contrast, our work seeks to balance adaptivity and efficiency through reusable workflow portfolios, enabling query-aware behavior without incurring the substantial overhead of fully dynamic workflow generation.

\section{Prompts for DiverseFlow}

\begin{tcolorbox}[
    breakable,              
    enhanced,
    colback=white,
    colframe=colorphase1,
    colbacktitle=colorphase1,
    coltitle=white,
    title=Performace-oriented Warm-up,
    sharp corners,
    boxrule=1pt,
    drop shadow
]
\begin{lstlisting}[basicstyle=\ttfamily\scriptsize]
You are building an LLM Graph (Python workflow) and corresponding Prompt to jointly solve {type} problems.
{dataset_description}
Referring to the given graph and prompt below, please reconstruct and optimize them. You can add, modify, or delete nodes, parameters, or prompts. Include your single modification in XML tags in your reply.

When optimizing, you can incorporate critical thinking methods like review, revise, ensemble (generating multiple answers through different/similar prompts, then voting/integrating/checking the majority to obtain a final answer), selfAsk, etc. Consider Python's loops (for) and conditional statements (if-elif-else).
Considering information loss, complex graphs may yield better results, but insufficient information transmission can omit the solution. It's crucial to include necessary context during the process.
Introducing multiple operators at appropriate points can enhance performance. If some provided operators are not yet used, try incorporating them.

## Operator Initialization and Usage (MUST follow exactly)

All operators MUST be initialized with `self.llm` in `__init__`. Example:
```
        self.custom = operator.Custom(self.llm)
        self.sc_ensemble = operator.ScEnsemble(self.llm)
        self.answer_generate = operator.AnswerGenerate(self.llm)
        self.custom_code_generate = operator.CustomCodeGenerate(self.llm)
        self.test = operator.Test(self.llm, dataset=self.dataset)
        self.programmer = operator.Programmer(self.llm)
```

Operator call formats (MUST follow exactly):

1. **Custom**
   Format: `custom(input: str, instruction: str) -> str`
   Example: `solution = await self.custom(input=problem, instruction=prompt_custom.SOLVE_PROMPT)`
   Note: Returns the response string directly. The input and instruction are directly concatenated (instruction+input). Placeholders are not supported. You MUST define the prompt variable (e.g., `SOLVE_PROMPT = "..."`) in the `<prompt>` section for every `prompt_custom.XXXX` you use in graph.

2. **ScEnsemble**
   Format: `sc_ensemble(solutions: List[str], problem: str) -> str`
   Example: `best = await self.sc_ensemble(solutions=[sol1, sol2, sol3], problem=problem)`
   Note: Returns the best solution string directly. Selects the best from multiple candidates via voting.

3. **AnswerGenerate**
   Format: `answer_generate(input: str) -> str`
   Example: `answer = await self.answer_generate(input=problem)`
   Note: Returns the answer string directly. Generates step-by-step reasoning internally.

4. **CustomCodeGenerate**
   Format: `custom_code_generate(problem: str, entry_point: str, instruction: str) -> str`
   Example: `code = await self.custom_code_generate(problem=problem, entry_point=entry_point, instruction="Analyze step by step and generate code.")`
   Note: Returns the code string directly. The instruction should encourage step-by-step thinking.

5. **Test**
   Format: `test(problem: str, solution: str, entry_point: str) -> dict with keys 'result' (bool) and 'solution' (str)`
   Example: `test_result = await self.test(problem=problem, solution=solution, entry_point=entry_point)`
   Note: Modify the input solution solution with public test cases. Always return test_result['solution'] (the improved solution). Use test_result['result'] (bool) for conditional retry if needed.

6. **Programmer**
   Format: `programmer(problem: str, analysis: str = 'None') -> str`
   Example: `result = await self.programmer(problem=problem, analysis="Step by step analysis")`
   Note: Writes and executes Python code, returns a string with the execution result.

Here is a graph and the corresponding prompt (prompt only related to the custom method) that performed excellently in a previous iteration (maximum score is 1). You must make further optimizations and improvements based on this graph.

<sample>
    <experience>{experience}</experience>
    <modification>(such as: add / delete / modify / ...)</modification>
    <score>{score}</score>
    <graph>{graph}</graph>
    <prompt>{prompt}</prompt>(only prompt_custom)
</sample>

Below are the logs of some results with the aforementioned Graph that performed well but encountered errors, which can be used as references for optimization:
{log}

Now, provide your optimization. The modified graph must differ from the provided example, and the specific differences should be noted within the <modification>xxx</modification> section.

**OPERATOR RULES (MUST follow strictly):**
1. Do NOT create new operators. Only use the operators listed in the Operator Usage section above.
2. All operators MUST be initialized in `__init__` with `self.llm` as the first argument. Example: `self.custom = operator.Custom(self.llm)`. Never call `operator.XXX()` without `self.llm`.
3. Follow the exact call format for each operator as specified in the Operator Usage section.
4. Loop iteration MUST <= 5 to avoid timeout.
5. The graph complexity should not exceed 8 nodes.

**`__call__` SIGNATURE AND RETURN FORMAT (MUST follow exactly):**
The `__call__` method signature MUST be:
```python
async def __call__(self, problem: str):
    # ... your workflow logic ...
    return solution, self.llm.usage_tracker.get_summary()["total_cost"]
```
- Input: `problem` (str) -- the math problem text.
- Return: a **2-tuple** of `(solution_string, cost_float)`. Do NOT return only the solution without cost. Do NOT return a 3-tuple or any other format.

**PROMPT GENERATION RULES (CRITICAL):**
- Only generate prompts used by Custom operator via `prompt_custom.YOUR_PROMPT_NAME` in the graph.
- You MUST define every `prompt_custom.XXXX` variable you reference in graph in the `<prompt>` section. For example, if you use `prompt_custom.SOLVE_PROMPT` in graph, you must output `SOLVE_PROMPT = "..."` in `<prompt>`.
- Other operators (ScEnsemble, Programmer, AnswerGenerate, CustomCodeGenerate, Test) have built-in prompts -- do NOT generate prompts for them.
- **The generated prompt MUST be plain text. Do NOT use placeholders like {{problem}}, {{entry_point}}, or {{input}}. The prompt will be directly concatenated with the input, NOT formatted with .format().**
- Remove any unused prompts from prompt_custom.

**MODIFICATION RULES:**
- **Only ONE detail point can be modified at a time**, and no more than 5 lines of code (including graph and prompt) may be changed per modification -- extensive modifications are strictly prohibited to maintain project focus!
- When introducing new functionalities in the graph, please make sure to import the necessary libraries or modules yourself, except for operator, prompt_custom, create_llm_instance, and CostManage, which have already been automatically imported.
- **Under no circumstances should Graph output None for any field.**
- Use custom methods to restrict your output format, rather than using code (outside of the code, the system will extract answers based on certain rules and score them).
- It is very important to format the Graph output answers, you can refer to the standard answer format in the log.
\end{lstlisting}

\end{tcolorbox}

\begin{tcolorbox}[
    breakable,              
    enhanced,
    colback=white,
    colframe=colorphase2,
    colbacktitle=colorphase2,
    coltitle=white,
    title=Complementarity-oriented Expansion,
    sharp corners,
    boxrule=1pt,
    drop shadow
]
\begin{lstlisting}[basicstyle=\ttfamily\scriptsize]
You are building an LLM Graph (Python workflow) and corresponding Prompt to jointly solve {type} problems.
{dataset_description}
Referring to the given graph and prompt below, please reconstruct and optimize them. You can add, modify, or delete nodes, parameters, or prompts. Include your single modification in XML tags in your reply.

When optimizing, you can incorporate critical thinking methods like review, revise, ensemble (generating multiple answers through different/similar prompts, then voting/integrating/checking the majority to obtain a final answer), selfAsk, etc. Consider Python's loops (for) and conditional statements (if-elif-else).
Considering information loss, complex graphs may yield better results, but insufficient information transmission can omit the solution. It's crucial to include necessary context during the process.
Introducing multiple operators at appropriate points can enhance performance. If some provided operators are not yet used, try incorporating them.

## Operator Initialization and Usage (MUST follow exactly)

All operators MUST be initialized with `self.llm` in `__init__`. Example:
```
        self.custom = operator.Custom(self.llm)
        self.sc_ensemble = operator.ScEnsemble(self.llm)
        self.answer_generate = operator.AnswerGenerate(self.llm)
        self.custom_code_generate = operator.CustomCodeGenerate(self.llm)
        self.test = operator.Test(self.llm, dataset=self.dataset)
        self.programmer = operator.Programmer(self.llm)
```

Operator call formats (MUST follow exactly):

1. **Custom**
   Format: `custom(input: str, instruction: str) -> str`
   Example: `solution = await self.custom(input=problem, instruction=prompt_custom.SOLVE_PROMPT)`
   Note: Returns the response string directly. The input and instruction are directly concatenated (instruction+input). Placeholders are not supported. You MUST define the prompt variable (e.g., `SOLVE_PROMPT = "..."`) in the `<prompt>` section for every `prompt_custom.XXXX` you use in graph.

2. **ScEnsemble**
   Format: `sc_ensemble(solutions: List[str], problem: str) -> str`
   Example: `best = await self.sc_ensemble(solutions=[sol1, sol2, sol3], problem=problem)`
   Note: Returns the best solution string directly. Selects the best from multiple candidates via voting.

3. **AnswerGenerate**
   Format: `answer_generate(input: str) -> str`
   Example: `answer = await self.answer_generate(input=problem)`
   Note: Returns the answer string directly. Generates step-by-step reasoning internally.

4. **CustomCodeGenerate**
   Format: `custom_code_generate(problem: str, entry_point: str, instruction: str) -> str`
   Example: `code = await self.custom_code_generate(problem=problem, entry_point=entry_point, instruction="Analyze step by step and generate code.")`
   Note: Returns the code string directly. The instruction should encourage step-by-step thinking.

5. **Test**
   Format: `test(problem: str, solution: str, entry_point: str) -> dict with keys 'result' (bool) and 'solution' (str)`
   Example: `test_result = await self.test(problem=problem, solution=solution, entry_point=entry_point)`
   Note: Modify the input solution solution with public test cases. Always return test_result['solution'] (the improved solution). Use test_result['result'] (bool) for conditional retry if needed.

6. **Programmer**
   Format: `programmer(problem: str, analysis: str = 'None') -> str`
   Example: `result = await self.programmer(problem=problem, analysis="Step by step analysis")`
   Note: Writes and executes Python code, returns a string with the execution result.

Here is a graph and the corresponding prompt (prompt only related to the custom method) that performed excellently in a previous iteration (maximum score is 1). You must make further optimizations and improvements based on this graph.

<sample>
    <experience>{experience}</experience>
    <modification>(such as: add / delete / modify / ...)</modification>
    <score>{score}</score>
    <graph>{graph}</graph>
    <prompt>{prompt}</prompt>(only prompt_custom)
</sample>

Below are the logs of some results with the aforementioned Graph that performed well but encountered errors, which can be used as references for optimization:
{log}

**SCORING**: Your workflow is evaluated by WEIGHTED accuracy --- solving under-covered queries counts much more than solving well-covered ones. The score shown above reflects this weighted metric, not raw accuracy.

**Query weight distribution**:
- {n_weight_high} queries covered by 0 workflows (weight=1.00, highest impact on your score)
- {n_weight_mid} queries covered by 1-3 workflows (weight=0.25-0.50)
- {n_weight_low} queries covered by 4+ workflows (weight<0.20, low impact)

**High-weight examples** (solving these types has the most impact on your score):
{high_weight_examples}

**Existing strengths** (avoid duplicating):
{existing_strengths}

**The workflow is pure Python --- use the full language freely.** Go beyond simple operator chains. Consider these design patterns:
- **Majority Vote**: Loop N calls + `collections.Counter` on extracted answers (more robust than ScEnsemble)
- **Dynamic Prompts**: Use `.format()` to inject prior outputs into prompts (define templates with `{placeholders}` in `<prompt>`)
- **Multi-Agent Debate**: Proposer -> Critic -> Judge, each seeing prior output
- **Conditional Routing**: Classify problem type -> if/else -> specialist prompt
- **Helper Methods**: `@staticmethod` for regex answer extraction, post-processing
- **Multi-Paradigm**: Combine forward reasoning + backward reasoning + code execution, then ensemble

You can import `re`, `collections`, `json`, `math`, etc. Think structurally different, not just prompt tweaks.

**Relaxed rules for diversity mode** (override the rules below where they conflict):
- Prompts MAY use `{placeholders}` filled via `.format()` in graph code (for dynamic prompt composition)
- Up to **10 lines** of code may be changed (structural changes need more room)
- You MAY define `@staticmethod` or helper methods on the Workflow class

Now, provide your optimization. The modified graph must differ from the provided example, and the specific differences should be noted within the <modification>xxx</modification> section.

**OPERATOR RULES (MUST follow strictly):**
1. Do NOT create new operators. Only use the operators listed in the Operator Usage section above.
2. All operators MUST be initialized in `__init__` with `self.llm` as the first argument. Example: `self.custom = operator.Custom(self.llm)`. Never call `operator.XXX()` without `self.llm`.
3. Follow the exact call format for each operator as specified in the Operator Usage section.
4. Loop iteration MUST <= 5 to avoid timeout.
5. The graph complexity should not exceed 8 nodes.

**`__call__` SIGNATURE AND RETURN FORMAT (MUST follow exactly):**
The `__call__` method signature MUST be:
```python
async def __call__(self, problem: str):
    # ... your workflow logic ...
    return solution, self.llm.usage_tracker.get_summary()["total_cost"]
```
- Input: `problem` (str) --- the math problem text.
- Return: a **2-tuple** of `(solution_string, cost_float)`. Do NOT return only the solution without cost. Do NOT return a 3-tuple or any other format.

**PROMPT GENERATION RULES (CRITICAL):**
- Only generate prompts used by Custom operator via `prompt_custom.YOUR_PROMPT_NAME` in the graph.
- You MUST define every `prompt_custom.XXXX` variable you reference in graph in the `<prompt>` section. For example, if you use `prompt_custom.SOLVE_PROMPT` in graph, you must output `SOLVE_PROMPT = "..."` in `<prompt>`.
- Other operators (ScEnsemble, Programmer, AnswerGenerate, CustomCodeGenerate, Test) have built-in prompts --- do NOT generate prompts for them.
- **The generated prompt MUST be plain text. Do NOT use placeholders like {{problem}}, {{entry_point}}, or {{input}}. The prompt will be directly concatenated with the input, NOT formatted with .format().**
- Remove any unused prompts from prompt_custom.

**MODIFICATION RULES:**
- **Only ONE detail point can be modified at a time**, and no more than 5 lines of code (including graph and prompt) may be changed per modification --- extensive modifications are strictly prohibited to maintain project focus!
- When introducing new functionalities in the graph, please make sure to import the necessary libraries or modules yourself, except for operator, prompt_custom, create_llm_instance, and CostManage, which have already been automatically imported.
- **Under no circumstances should Graph output None for any field.**
- Use custom methods to restrict your output format, rather than using code (outside of the code, the system will extract answers based on certain rules and score them).
- It is very important to format the Graph output answers, you can refer to the standard answer format in the log.
\end{lstlisting}

\end{tcolorbox}
\appendix
\end{document}